\documentclass[10pt,journal,compsoc]{IEEEtran}

\usepackage{graphicx}
\usepackage{amsmath, bm}
\usepackage{amssymb}
\usepackage{tabularx}
\usepackage[linesnumbered,vlined,ruled]{algorithm2e}
\usepackage[greek,english]{babel}
\usepackage{multicol,multirow,booktabs,mathabx}
\usepackage{makecell}
\usepackage{array}
\usepackage{float}
\usepackage{booktabs}
\usepackage[table]{xcolor}
\usepackage{pifont}
\usepackage{dblfloatfix}
\newcommand{\xmark}{\ding{55}}%
\newcolumntype{P}[1]{>{\centering\arraybackslash}p{#1}}

\usepackage{arydshln}
\usepackage{setspace}
\usepackage[colorlinks=true, urlcolor=blue, linkcolor=red]{hyperref}

\ifCLASSOPTIONcompsoc
  \usepackage[nocompress]{cite}
\else
  \usepackage{cite}
\fi

\ifCLASSINFOpdf
\else
\fi

\usepackage{arydshln}

\begin{document}

\title{Track Initialization and Re-Identification for~3D Multi-View Multi-Object Tracking}

\author{Linh~Van~Ma, Tran~Thien~Dat~Nguyen, Ba-Ngu~Vo, Hyunsung~Jang, Moongu~Jeon% <-this % stops a space
\IEEEcompsocitemizethanks{\IEEEcompsocthanksitem Linh Van Ma and Moongu Jeon are with the School of Electrical Engineering and Computer Science at GIST, Gwangju, Korea (e-mail: \{linh.mavan, mgjeon\}@gist.ac.kr).%\protect\\
% E-mail: see http://www.michaelshell.org/contact.html
\IEEEcompsocthanksitem Tran Thien Dat Nguyen and Ba-Ngu Vo are with the School of Electrical Engineering, Computing and Mathematical Sciences, Curtin University, Australia (e-mail: \{t.nguyen1, ba-ngu.vo\}@curtin.edu.au).
\IEEEcompsocthanksitem Hyunsung~Jang is with the Department of EO/IR Systems Research and Development, LIG Nex1, Korea (e-mail: hyunsung.jang@lignex1.com).}}

% \markboth{Journal of \LaTeX\ Class Files,~Vol.~14, No.~8, August~2015}%
% {Shell \MakeLowercase{\textit{et al.}}: Bare Demo of IEEEtran.cls for Computer Society Journals}

\IEEEtitleabstractindextext{%
\begin{abstract}
We propose a 3D multi-object tracking (MOT) solution using only 2D detections from monocular cameras, which automatically initiates/terminates tracks as well as resolves track appearance-reappearance and occlusions. Moreover, this approach does not require detector retraining when cameras are reconfigured but only the camera matrices of reconfigured cameras need to be updated. Our approach is based on a Bayesian multi-object formulation that integrates track initiation/termination, re-identification, occlusion handling, and data association into a single Bayes filtering recursion. However, the exact filter that utilizes all these functionalities is numerically intractable due to the exponentially growing number of terms in the (multi-object) filtering density, while existing approximations trade-off some of these functionalities for speed. To this end, we develop a more efficient approximation suitable for online MOT by incorporating object features and kinematics into the measurement model, which improves data association and subsequently reduces the number of terms. Specifically, we exploit the 2D detections and extracted features from multiple cameras to provide a better approximation of the multi-object filtering density to realize the track initiation/termination and re-identification functionalities. Further, incorporating a tractable geometric occlusion model based on 2D projections of 3D objects on the camera planes realizes the occlusion handling functionality of the filter. Evaluation of the proposed solution on challenging datasets demonstrates significant improvements and robustness when camera configurations change on-the-fly, compared to existing multi-view MOT solutions. The source code is publicly available at \href{https://github.com/linh-gist/mv-glmb-ab}{https://github.com/linh-gist/mv-glmb-ab}.
\end{abstract}

\begin{IEEEkeywords}
Multi-view, Multi-sensor, Multi-object Visual Tracking, Occlusion Handling, Generalized Labeled Multi-Bernoulli, Re-Identification, Adaptive Birth.
\end{IEEEkeywords}}

\maketitle

\IEEEdisplaynontitleabstractindextext

% \fi
\IEEEpeerreviewmaketitle

%%%%%%%%%%%% SECTION %%%%%%%%%%%%%%%
\section{Introduction}\label{sec:introduction}

Visual tracking is a branch of multi-object tracking (MOT), which aims at estimating an unknown number of object trajectories from video sequences. There are two main approaches to MOT: track-by-detection and track-before-detect. In the former, object detection is obtained independently and then supplied to the tracker to generate track estimates, while the latter operates on the input signal without object detection. In practice, track-before-detect is computationally intensive and track-by-detection is more commonly used, especially for visual MOT due to the efficiency and reliability of 2D object detectors. The main challenges are the uncertainties in the number of objects and data association. Numerous (track-by-detection) MOT algorithms have been developed, usually under the three main paradigms: multiple hypothesis tracking (MHT) \cite{thomaidis2013multiple}; joint probabilistic data association (JPDA) \cite{blackman1999design}; and random finite set (RFS) \cite{ristic2016overview}.

The advancement and popularity of 2D visual MOT is mainly driven by fast and reliable 2D object detectors. When object motion is slow (relative to the frame rate) and object detection is accurate, simple trackers with kinematic/shape cues such as SORT \cite{wojke2017simple} and IoU-Tracker \cite{bochinski2017high} can achieve accurate tracking rate with little computation time. For challenging scenarios, with higher levels of uncertainty, more sophisticated trackers are needed \cite{kim2019labeled,nguyen2021tracking}. In addition, objects in 2D images are usually rich in visual features (e.g., pedestrians walking on the streets) and visual cues that can be exploited to distinguish different objects \cite{liang2022rethinking,zhang2021fairmot}, improve data association as well as re-identification of lost tracks when they re-appear \cite{liang2022rethinking}, assuming slow variations in the visual appearance of objects.

Since objects such as people, cars, drones, etc. reside in the 3D world, 2D trajectories are not adequate for scene understanding or post-tracking analysis \cite{bridgeman2019multi,bradler2021urban}, which requires 3D visual tracking. Moreover, trajectories in 3D world frame are more informative for applications such as sports analytics, age care, school environment monitoring, etc. Multi-view data also helps resolving occlusions since objects occluded in one view can be detected in other views. 

A popular solution to 3D visual tracking is applying MOT to 3D detections obtained by using multi-view fusion to reconstruct objects in 3D from the 2D multi-view detections \cite{chavdarova2017deep,chavdarova2018wildtrack}. However, unlike the detection of objects in 2D images, determining the 3D locations of objects from multi-view images is challenging \cite{ning2024dilf,lupion20243d}. While some deep learning solutions can achieve high detection accuracy, training 3D object detectors is computationally demanding, especially for high dimensional scenarios (e.g., large number of cameras) \cite{baque2017deep}. Moreover, when the camera configurations change, the detectors need to be retrained, which limits the online operation of the tracker.   

\begin{figure}[tbh]
\centering
\includegraphics[width=0.99\columnwidth]{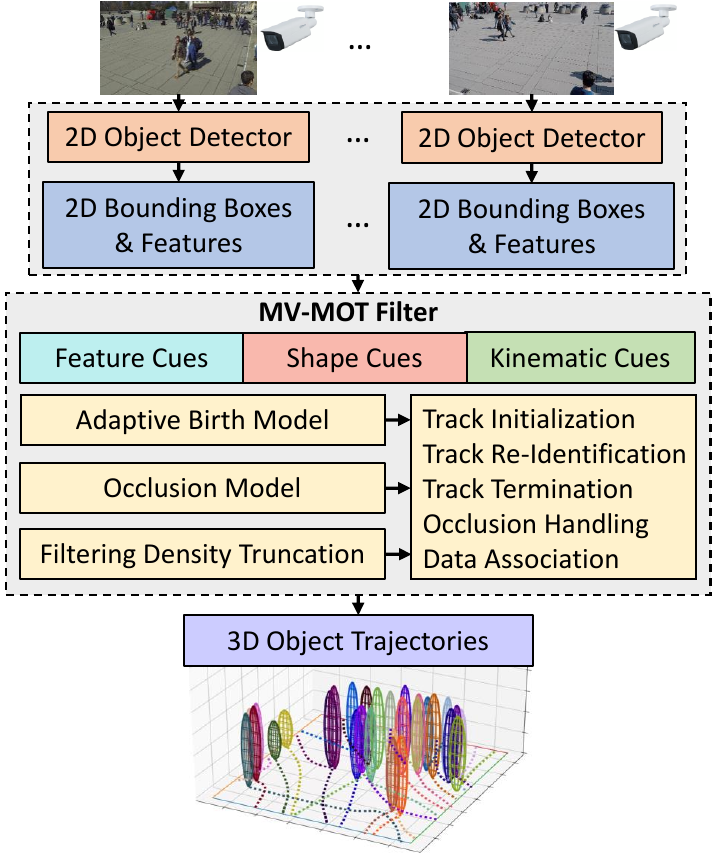}
\caption{Schematic of the proposed 3D MV-MOT solution. Multi-view detections (bounding boxes and visual features from all cameras) is supplied to the MV-MOT filter, which integrates multi-object dynamic and measurement models to realize all MOT functionalities.\label{fig:overview}}
\end{figure}

We propose a 3D visual tracking algorithm that exploits the extracted features from 2D multi-view detections via multi-sensor MOT to automatically initiates/terminates and re-identifies tracks as well resolving occlusions. Unlike many of the 3D visual tracking techniques that only provide global trajectories on the ground plane, the proposed solution processes 2D detections from multiple monocular cameras, online, to provide trajectories in 3D world frame. Our approach takes advantage of advances in 2D object detection and multi-sensor MOT that exploits geometric information from cameras with overlapping fields of view to accurately estimate the shape and position of 3D objects.  The proposed multi-view MOT (MV-MOT) algorithm has a linear complexity in the number of detections across all cameras. Moreover, it does not require detector to be retrained when the cameras are reconfigured, and is amenable to seamless fusion with other types of sensor data. Performance evaluations on challenging datasets demonstrate significant improvements in tracking accuracy compared to existing solutions, and robustness when camera configurations change on-the-fly. Ablation studies are also presented to illustrate its advantages. A schematic of the proposed 3D visual tracking solution is shown in Fig. \ref{fig:overview}. Our contributions are summarized as follows: 

\textbullet{} Novel multi-object dynamic and measurement models that jointly account for object kinematics, shapes, visual features on different cameras, and occlusion (including partial and complete occlusion);   

\textbullet{} An approximation of the MV-MOT filter that automatically performs 3D track initialization/termination, re-identification, and occlusion handling using 2D multi-view monocular detection, with linear complexity in the number of detections across the cameras. 

\textbullet{} Extensive experiments to evaluate the performance on challenging benchmarks including the Curtin multi-camera (CMC) \cite{ong2020bayesian} and WILDTRACK (WT) \cite{chavdarova2018wildtrack} datasets. 

The paper is organized as follows. In Section \ref{sec:backgrounds} related works in 2D/3D object detection and tracking are discussed. Section \ref{sec:MV-GLMB} introduces the dynamic and measurement models together with the Bayes recursion that form our 3D visual MV-MOT solution. In Section \ref{sec:implement}, we propose an efficient approximation of the MV-MOT filter that realizes automatic track initiation/re-identification and occlusion resolution. Extensive experiments to verify the effectiveness of our tracking solutions are given in Section \ref{sec:experiment}, and Section \ref{sec:conclusion} concludes the paper.

%%%%%%%%%%%% SECTION %%%%%%%%%%%%%%%

\section{Related Works}\label{sec:backgrounds}
\subsection{Visual Multi-Object Detection}
Multi-object detection from 2D images is a key research topic in computer vision. Early detectors use template matching to localize objects in images  \cite{betke1995fast}. Many learning-based solutions rely on trainable classifiers such as support vector machines or Adaboost to detect objects \cite{viola2005detecting} using features such as Haar, scale-invariant feature transform (SIFT) \cite{lowe2004distinctive}, and histogram of oriented gradients (HOG) \cite{dalal2005histograms}. Deep learning has become popular in object detection due to the utility of convolutional neural networks (CNNs) \cite{girshick2014rich}. Combinations of effective region-proposal algorithms \cite{zitnick2014edge} and CNN features (i.e., features extracted from CNNs) have resulted in real-time, high-performance 2D object detectors \cite{ren2015faster}. YOLO algorithms that bypass the region-proposal step by casting detection as a regression problem are significantly more efficient \cite{redmon2018yolov3}. Recently, algorithms that formulate object detection as a set of learning tasks have also been proposed \cite{carion2020end}. Publicly available large-scale datasets have been instrumental for the fast-paced development of 2D object detection solutions, especially learning-based methods \cite{lin2014microsoft,russakovsky2015imagenet}.

Detecting occluded objects in 2D images is a challenging problem. Multi-view images provide more accurate detection than single-view images by fusing information from different views. In \cite{fleuret2007multicamera}, a probabilistic occupancy map is constructed from background-subtracted images to locate objects on the ground plane. However, this method tends to generate a high number of false alarms. Nonetheless, this could be reduced by the Bayesian network-based technique proposed in \cite{peng2015robust}. Alternatively, Gibbs sampling is used in \cite{ge2010crowd} to generate the number of objects and their spatial locations from a posterior (conditioned on 2D detections). CNN features can also be used for multi-view detection, e.g., in \cite{baque2017deep} a discriminative CNN feature extraction module is used in conjunction with a generative occlusion model to construct an existence probability map of objects on the ground plane, while in \cite{hou2020multiview}, CNN features are projected to the ground plane and then fed into classifiers to localize objects. Methods based on similar projections are also proposed in \cite{zhang2021cross,song2021stacked}.

\subsection{Visual Multi-Object Tracking}
Visual MOT solutions can be categorized as online or batch. Batch algorithms estimate object trajectories from a batch of data, with computational complexities per time step growing over the time window. On the other hand, online algorithms estimate object trajectories at each time step when new data arrive, with computational complexities per time step that are independent of time, and hence are preferred over batch algorithms in practice. In 2D visual MOT, algorithms that exploit only motion and shape information are fast but cannot handle complex scenarios \cite{bochinski2017high,wojke2017simple}. Alternatively, visual features of objects can be used to improve tracking. Hand-crafted features (e.g., SIFT, HOG) are not effective in distinguishing different objects \cite{chen2018real}, and CNN features are more suitable due to the multi-scale representation. In \cite{chen2018real} separate models for detection and feature extraction are used. While using a single model for both tasks was demonstrated to have better efficiency in \cite{wang2020towards}, a balance between the two tasks needs to be considered \cite{zhang2021fairmot}. State-of-the-art (SOTA) 2D multi-object trackers that utilize feature cues in the literature include POI \cite{yu2016poi}, MOTDT \cite{chen2018real}, DeepSORT \cite{wojke2018deep}, and GSDT \cite{wang2021joint}.

Multi-view MOT solutions are becoming increasingly important due to the proliferation of cameras and the better tracking performance over single-view techniques. Homography constraints are used to track human feet in \cite{khan2006multiview}, while in \cite{eshel2008homography} heads are localized in single-view images and then transformed to world coordinates to perform tracking. In \cite{hu2006principal}, principal axes are used to associate tracklets between cameras, while in \cite{xu2017cross} advanced semantic cues are used. In \cite{xu2016multi}, 2D detections are mapped into 3D positions, and then combined with relevant cues (i.e., motion, features, geometry proximity) to associate tracklets using a hierarchical composition model. In \cite{zhang2022mutr3d, pang2023standing}, 3D objects of different classes are tracked with a 3D track query model using multiple monocular cameras for autonomous driving applications.

Occlusion handling is an important functionality of visual tracking. In the single-view case, certain solutions rely on detectors that can localize parts of the objects \cite{Ouyang2018jointly}, albeit training such detectors to yield accurate localization results is difficult. A popular approach is to use designated modules that analyze occlusion, using object depth \cite{Ma2010DepthAO}, or spatial information of objects and their interactions to resolve occlusion \cite{Stadler2021improving,Xiaoding2020robust}. In the multi-view case, occlusion can also be implicitly resolved in the multi-view data fusion process, usually exploiting object locations, either at the detection or tracking step \cite{baque2017deep,ong2020bayesian}.  

Data association is a crucial and challenging problem in track-by-detection MOT. Simple algorithms such as the global nearest neighbor (GNN) \cite{blackman1999design} consider a single hypothesis of data association. More sophisticated MOT frameworks such as MHT, JPDA, and RFS have demonstrated improved tracking performance by keeping multiple data association hypotheses. The labeled RFS solutions \cite{vo2013labeled}, such as the GLMB filter, are well suited for online and multi-view MOT due to the low-complexity and efficiency \cite{vo2019multi}. Indeed, the GLMB filter has been used in various computer vision problems \cite{kim2019labeled,nguyen2021tracking}, including multi-sensor data association for multi-view occlusion handling \cite{ong2020bayesian}.

While MOT functionalities such as data association, track initiation/termination, re-identification, and occlusion handling are captured in the GLMB filtering recursion \cite{vo2019multi}, an exact implementation realizing all these functionalities is numerically intractable. In \cite{ong2020bayesian}, an approximation was developed to address occlusion, but the re-identification functionality was neglected, and track initiation requires a combination of accurate prior birth models (which is not always available) with clustering. While object features improve tracking performance \cite{liang2022rethinking,zhang2021fairmot}, they have not been exploited by the filter to improve data association and resolve track re-identification. Moreover, the occlusion model in \cite{ong2020bayesian} does not account for partial occlusions, and thus was unable to exploit SOTA 2D object detection technique \cite{wang2020robust}. Without accurate prior information, initializing tracks from multi-sensor measurements is challenging due to the unknown number of new tracks, miss-detection, false alarms, and the large number of possible detection combinations from multiple sensors. The recent solution in \cite{trezza2022multi} utilizes Monte Carlo (MC) technique to initialize tracks where existence probabilities depend on their measurement likelihood and how likely the detections are already associated with known tracks. While, this solution can be directly applied to visual tracking, it is difficult to find a balance between speed and accuracy not to mention the inability resolve track appearance-reappearance.

%%%%%%%%%%%% SECTION %%%%%%%%%%%%%%%
\section{Bayesian Multi-View MOT}\label{sec:MV-GLMB}
This section presents a Bayesian tracker that can handle all functionalities of a multi-view multi-object tracker from automatic track initialization, termination, re-identification to multi-view data association and occlusion handling. In particular, details on the object dynamic and measurement models will be given together with a Bayes recursion that propagates the multi-object density over time. Notations commonly used in this paper are tabulated in Tab. \ref{tbl:notationlist}. 

\begin{table}[!ht]
\centering
% \global\long\def\arraystretch{1.3}%
 \caption{List of symbols.\label{tbl:notationlist}}
 \footnotesize
\begin{tabular}{|c|>{\centering}p{6cm}|}
\hline 
Notation & Description\tabularnewline
\hline 
$\otimes$ & Kronecker product (for matrices)\tabularnewline
$h^{X}$ & $\prod\limits _{x\in X}h(x)$ with $h^{\emptyset}=1$\tabularnewline
$\langle f,g\rangle$ & $\int f(x)g(x)dx$, inner product of $f$ and $g$, \tabularnewline
$j:k$ & $j,j+1,...,k$\tabularnewline
$x^{(j:k)}$ & $x^{(j)},x^{(j+1)},...,x^{(k)}$\tabularnewline
$x_{j:k}$ & $x_{j},x_{j+1,}...,x_{k}$\tabularnewline
$\mathbb{X}$ & single object state space\tabularnewline
$\mathbb{L}$ & discrete label space\tabularnewline
$\mathbb{B}$ & discrete label space for new birth objects\tabularnewline
$\boldsymbol{X}$ & labled multi-object state\tabularnewline
$\mathcal{L}(\boldsymbol{X})$ & set of labels of multi-object state $\boldsymbol{X}$\tabularnewline
$\boldsymbol{x}=(x,\ell)$ & labeled single-object state (with label $\ell$)\tabularnewline
$\boldsymbol{\pi}$ & multi-object density\tabularnewline
$\Omega$ & MS/MV-GLMB recursion operator\tabularnewline
$\{(r_{B}^{(\ell)},p_{B}^{(\ell)})\}_{\ell\in\mathbb{B}}$ & parameters of new birth objects\tabularnewline
$f_{S,+}(x_{+}|x,\ell)$ & single-object transition density\tabularnewline
$P_{S,+}(\boldsymbol{x})$ & survival probability of labeled state $\boldsymbol{x}$\tabularnewline
$\mathbb{Z}^{(c)}$ & measurement space of camera $c$\tabularnewline
$Z^{(c)}$ & set of measurements of camera $c$\tabularnewline
$z^{(c)}$ & single-view measurement of camera $c$\tabularnewline
$g^{(c)}(z^{(c)}|\boldsymbol{x})$ & single-object single-view measurement likelihood function for camera
$c$\tabularnewline
$g^{(c)}(Z^{(c)}|\boldsymbol{X})$ & multi-object single-view measurement likelihood function for camera
$c$\tabularnewline
$\boldsymbol{g}\left(Z|\boldsymbol{X}\right)$ & multi-object multi-view measurement likelihood function\tabularnewline
$P_{D}^{(c)}(\boldsymbol{x},\boldsymbol{X})$ & detection probability for camera $c$\tabularnewline
$\alpha^{(\ell,c)}$ & observed feature of object $\ell$ at camera $c$\tabularnewline
$\Phi^{(c)}(x)$ & box bounding an object with state $x$ in camera $c$'s image plane\tabularnewline
$\gamma^{\left(c\right)}$ & association map for camera $c$\tabularnewline
$\gamma$ & multi-view association map \tabularnewline
$\Gamma^{\left(c\right)}$ & space of association maps for camera $c$\tabularnewline
$\Gamma$ & space of multi-view association maps \tabularnewline
$\mathcal{L}_{\gamma^{(c)}}$ & live label set of association map $\gamma^{(c)}$\tabularnewline
$\delta_{Y}[X]$ & generalized Kronecker delta function, takes on 1 if $X=Y$, and 0
otherwise\tabularnewline
$\mathcal{N}(.;\mu,P)$ & Gaussian pdf with mean $\mu$ and covariance $P$\tabularnewline
\hline 
\end{tabular}
\end{table}

\subsection{Object Dynamic Model}\label{subsec:MOT-transition}

The state $\boldsymbol{x}=(x,\ell)$ of an object consists of attribute $x$ from an attribute space $\mathbb{X}$ and a label $\ell$ from a discrete label space $\mathbb{L}$. An object born at time $k$, is assigned a time-invariant label $\ell=(k,\iota)$, where $\iota$ is a unique index to differentiate objects born at the same time. The attribute $x$ consists of 3D position $\zeta$, 3D velocity $\dot{\zeta}$, and shape parameter $\varsigma$. The \textit{multi-object state} at a given time $k$ is a \textit{finite set of individual object states} in $\mathbb{X}\times\mathbb{L}$ with distinct labels \cite{vo2013labeled}.   

At time $k$, a set (possibly
%possible 
empty) 
of new objects is born. The set of all possible labels of object born at time $k$ is a subset of $\mathbb{L}$, denoted by $\mathbb{B}$. A new object with label $\ell$ is born with probability $r_{B}^{(\ell)}$, and conditional on which its attribute is distributed according to $p_{B}^{(\ell)}$. The birth parameters $\{(r_{B}^{(\ell)},p_{B}^{(\ell)})\}_{\ell\in\mathbb{B}}$ could be provided apriori (if statistics of newborns are known), or estimated from the data. 

Given a multi-object state $\boldsymbol{X}$ at time $k$, each $(x,\ell)\in\boldsymbol{X}$ either survives to the next time with probability $P_{S,+}(x,\ell)$ or dies with probability $1-P_{S,+}(x,\ell)$. Conditional on survival the object takes on the new state $(x_{+},\ell_{+})$ according to the transition density $f_{S,+}(x_{+}|x,\ell)\delta_{\ell}[\ell_{+}]$ \cite{vo2013labeled}, where the generalized Kronecker delta $\delta_{\ell}[\ell_{+}]$, defined to be 1 when $\ell=\ell_{+}$ and 0 otherwise, ensures the label remains unchanged. The multi-object state $\boldsymbol{X}_{+}$ at the next time is the superposition of newborns and surviving objects, and is distributed according to the \textit{multi-object Markov transition density} $\boldsymbol{f}_{+}(\boldsymbol{X}_{+}|\boldsymbol{X})$ (an explicit expression is not needed in this work, nonetheless it can be found in \cite{vo2013labeled}). Hereon, we use the subscript '+' to indicate the next time.

In this work, we use the survival probability model proposed in \cite{kim2019labeled}. The shape parameter $\varsigma$ is a triplet of (logarithms of) the half-lengths of the principal axes of the ellipsoid containing the object, and follows a random-walk model. The kinematics ($\zeta$, $\dot{\zeta}$) follows a nearly constant velocity model. Specifically, given the current attribute $x$, the next attribute $x_{+}$ is distributed by \cite{ong2020bayesian}
\begin{equation}
f_{S,+}(x_{+}|x,\ell)=\mathcal{N}\left(x_{+};Fx+b,Q\right),\label{eq:transition_density}
\end{equation}
where 
\[
F=\left[\begin{array}{cc}
I_{3}(T) & 0_{6\times3}\\
0_{3\times6} & I_{3}
\end{array}\right],I_{3}(T)=I_{3}\otimes\left[\begin{array}{cc}
1 & T\\
0 & 1
\end{array}\right],
\]
\[
b=\left[\begin{array}{c}
0_{6\times1}\\
-\upsilon^{(\varsigma)}/2
\end{array}\right],Q=\left[\begin{array}{cc}
V(\upsilon^{(\zeta)},T) & 0_{6\times3}\\
0_{3\times6} & \textrm{diag}(\upsilon^{(\varsigma)})
\end{array}\right],
\]
\[
V(\upsilon^{(\zeta)},T)=\textrm{diag}(\upsilon^{(\zeta)})\otimes\left[\begin{array}{c}
\frac{T^{2}}{2}\\
T
\end{array}\right]\left[\begin{array}{cc}
\frac{T^{2}}{2} & T\end{array}\right],
\]
$T$ is the sampling period, $\upsilon^{(\zeta)}$ and $\upsilon^{(\varsigma)}$ are 3D vectors of noise variances for the position and shape parameter, respectively. The Gaussianity of the logarithms of the half-lengths ensures that they are non-negative. This is equivalent to log-normal distributions of these half-lengths with unit-mean and variances $e^{\upsilon_{i}^{(\varsigma)}}-1$, $i=1,2,3$ \cite{ong2020bayesian}.

\subsection{Multi-View Measurement Model\label{subsec:MOT-observation}}

Given cameras $1,...,C$ and a multi-object state $\boldsymbol{X}$, each $\boldsymbol{x}\in\boldsymbol{X}$ is detected by camera $c$ with probability $P_{D}^{\left(c\right)}\left(\boldsymbol{x};\boldsymbol{X}\right)$ and generates the single-view measurement $z^{\left(c\right)}\in\mathbb{Z}^{(c)}$ ($\mathbb{Z}^{(c)}$ is the measurement space of the camera $c$) with likelihood $g^{\left(c\right)}(z^{\left(c\right)}|\boldsymbol{x})$, or miss-detected with probability $1-P_{D}^{\left(c\right)}\left(\boldsymbol{x};\boldsymbol{X}\right)$. While the detection probability is assumed independent of the other (or all) objects in most MOT algorithms, this assumption is not valid in occlusions. The objects in $\boldsymbol{X}\backslash\{\boldsymbol{x}\}$ could occlude $\boldsymbol{x}$, which translates to a low detection probability for $\boldsymbol{x}$. Thus a suitable detection probability model that accounts for occlusion is needed for occlusion handling \cite{ong2020bayesian}.

\subsubsection{Single-View Single-Object Measurement Model}

Conditional on detection by camera $c$, $\boldsymbol{x}$ is observed as a 2D bounding box and a feature vector, i.e., $z^{(c)}=(z_{p}^{(c)},z_{e}^{(c)},z_{f}^{(c)})$ where $z_{p}^{(c)}$ is the box center, $z_{e}^{(c)}$ is its extent (parameterized by the logarithms of the width and height in the camera $c$'s image plane), and $z_{f}^{(c)}$ is the feature vector (pertaining to appearance or identity). Since the kinematic and feature observations of an object are independent, the single-view single-object measurement likelihood $g^{\left(c\right)}(z^{\left(c\right)}|\boldsymbol{x})$ can be written as
\begin{equation}
g^{(c)}(z_{p}^{(c)},z_{e}^{(c)},z_{f}^{(c)}|\boldsymbol{x})=g_{b}^{(c)}(z_{p}^{(c)},z_{e}^{(c)}|x,\ell)g_{f}^{(c)}(z_{f}^{(c)}|\ell),\label{eq:solikelihood-whole}
\end{equation}
where $g_{b}^{(c)}$ and $g_{f}^{(c)}$ are, respectively, the bounding box and feature measurement likelihoods.  

The bounding box measurement $(z_{p}^{(c)},z_{e}^{(c)})$ is a noisy version of the box $\Phi^{(c)}(x)$ bounding the image of object $(x,\ell)$ in camera $c$'s image plane, which can be computed analytically via the projection matrix, see \cite{zhang2000flexible}. Hence, likelihood of $(z_{p}^{(c)},z_{e}^{(c)})$ is given by \cite{ong2020bayesian}
\begin{equation}
g_{b}^{(c)}(z_{p}^{(c)},z_{e}^{(c)}|x,\ell)=
\mathcal{N}\left(\left[\begin{array}{c}
z_{p}^{(c)}\\
z_{e}^{(c)}
\end{array}\right];\Phi^{(c)}(x),\textrm{diag}\left(\left[\begin{array}{c}
\upsilon_{p}^{(c)}\\
\upsilon_{e}^{(c)}
\end{array}\right]\right)\right),\label{likelihood_eq}
\end{equation}
where $\upsilon_{p}^{(c)}$ and $\upsilon_{e}^{(c)}$ are the noise variances for the center and the extent (in logarithm) of the box, respectively.

The feature measurement vector $z_{f}^{(c)}$ captures the object's visual appearance, e.g., color histograms, HSV features, Deep Learning features. Visual features can be used to identify objects since they are relatively stable \cite{zhang2021fairmot} or slowly varying with time \cite{wang2020towards}. Nonetheless, visual features can suddenly change \cite{wojke2017simple}, and are not always reliable \cite{zhang2021fairmot}. Thus, visual features models usually accommodate several modes of observation \cite{wojke2017simple}. Without loss of generality, we use a likelihood for $z_{f}^{(c)}$ with two modes, a strong mode to capture the stable slow variation, and a weaker mode to capture the sudden changes. Specifically, 
\begin{equation}
g_{f}^{(c)}(z_{f}^{(c)}|\ell)\thinspace\propto\thinspace\sigma s_{f}(z_{f}^{(c)},\alpha^{(\ell,c)})+\bar{\sigma}s_{f}(z_{f}^{(c)},\bar{\alpha}^{(\ell,c)}),\label{eq:sv-so-feature}
\end{equation}
where: $s_{f}$ is a non-negative function that monotonically increases with the similarity between its arguments; $\alpha^{(\ell,c)}$ and $\bar{\alpha}^{(\ell,c)}$ are, respectively, the nominal feature vectors for the stable and unstable modes, with respective weights $\sigma$ and $\bar{\sigma}$. Further, following \cite{wang2020towards}, the slow variation of the feature vector is modeled by adaptively updating the nominal feature at each time step via 
\[
\alpha_{+}^{(\ell,c)}=\vartheta_{0}\alpha^{(\ell,c)}+(1-\vartheta_{0})z_{f}^{(c)},
\]
where $\vartheta_{0}$ is a weight that controls the contribution of the observed data to the nominal feature. In essence, $\alpha^{(\ell,c)}$ is the exponential moving average of the observed feature with momentum $\vartheta_{0}$. The initial the feature $\alpha^{(\ell,c)}$, of object $\ell$ at camera $c$, can take on the feature computed from the measurement that it is initialized with or some prior value if it is initially misdetected.

\subsubsection{Single-View Multi-Object Measurement Model}\label{subsec:singlemeas}

The measurement set $Z^{(c)}$ from camera $c$ is a superposition of object-originated measurements and independent false positives (or clutter). Conditional on the multi-object state $\boldsymbol{X}$, the object-originated measurements are statistically independent \cite{vo2013labeled}. False positives are commonly parameterized by an intensity function $\kappa^{\left(c\right)}$, where the number of false positives is Poisson distributed with mean $\langle\kappa^{\left(c\right)},1\rangle$, and individual false positives are independent and identically distributed according to $\kappa^{\left(c\right)}/\langle\kappa^{\left(c\right)},1\rangle$, where $\langle f,g\rangle=\int f(x)g(x)dx$. In most MOT algorithms $\kappa^{\left(c\right)}$ is often assumed constant and known apriori. Nonetheless, it can also be estimated on-the-fly along with the multi-object state, albeit with additional computations \cite{do2022robust}. 

To account for unknown data association, it is necessary to consider different object-to-measurement mappings. At time $k$, an \textit{association map} for camera $c$ is a mapping $\gamma^{\left(c\right)}:\mathbb{L}\rightarrow\{-1:|Z^{\left(c\right)}|\}$ such that each label can only be mapped to at most one measurement, where $|Z^{\left(c\right)}|$ denotes the cardinality of $Z^{\left(c\right)}$ \cite{vo2013labeled}. For a label $\ell$, $\gamma^{\left(c\right)}(\ell)=-1$ represents a non-existent object, $\gamma^{\left(c\right)}(\ell)=0$ represents a miss-detection at camera $c$, while $\gamma^{\left(c\right)}(\ell)>0$ represents the scenario that $\ell$ generates measurement $z_{\gamma^{(c)}\left(\ell\right)}^{(c)}$ at camera $c$. Let $\Gamma^{\left(c\right)}$ denote the set of all association maps, $\mathcal{L}(\boldsymbol{X})$ the set of labels of multi-object state $\boldsymbol{X}$, and $\mathcal{L}_{\gamma^{(c)}}$ $\triangleq\left\{ \ell:\gamma^{(c)}\left(\ell\right)\geq0\right\} $ is the live label set of $\gamma^{(c)}$. Then, the \textit{single-view multi-object measurement likelihood} for camera $c$ is given by \cite{vo2013labeled}
\begin{equation}
\!\!\boldsymbol{g}^{(c)}(Z^{\left(c\right)}|\boldsymbol{X})\propto\!\sum_{\gamma^{\left(c\right)}\in\Gamma^{\left(c\right)}}\!\!\!\!\delta_{\mathcal{L}(\gamma^{\left(c\right)})}[\mathcal{L}\left(\boldsymbol{X}\right)]\!\left[\psi_{Z^{\left(c\right)},\boldsymbol{X}}^{(c,\gamma^{\left(c\right)}(\mathcal{L}(\cdot)))\!\!}\left(\cdot\right)\!\right]^{\boldsymbol{X}}\!\!,\label{e:Single_Sensor_Lkhd}
\end{equation}
where $\delta_{A}[B]=1$ if $A=B$ and zero otherwise, 
\begin{align}
\psi_{\{z_{1:|Z^{\left(c\right)}|}^{(c)}\},\boldsymbol{X}}^{(c,j)}\left(\boldsymbol{x}\right)= & \begin{cases}
1-P_{D}^{\left(c\right)}\left(\boldsymbol{x};\boldsymbol{X}\right), & \!\!\!\!\!j=0\\
\frac{P_{D}^{\left(c\right)}\left(\boldsymbol{x};\boldsymbol{X}\right)g^{\left(c\right)}(z_{j}^{(c)}|\boldsymbol{x})}{\kappa^{\left(c\right)}(z_{j}^{(c)})}, & \!\!\!\!\!j>0
\end{cases}.\label{e:Psi}
\end{align}

\subsubsection{Multi-View Multi-Object Measurement Model}

Noting that $\gamma^{(1)}(\ell)=...=\gamma^{(C)}(\ell)=-1$ if $\ell$ does not exist, we define a \textit{multi-view association map} as a tuple $\gamma\triangleq(\gamma^{\left(1:C\right)})$ of association maps such that, $\gamma^{(c)}(\ell)=-1$ for any $c$ implies, $\gamma^{(c)}(\ell)=-1$ for all $c$. This means $\gamma:\mathbb{L}\rightarrow\{-1\}^{C}\uplus(\mathbb{J}^{(1)}\times\cdots\times\mathbb{J}^{(C)})$, where $\mathbb{J}^{(c)}\triangleq\{0:\left|Z^{(c)}\right|\}$. Let $\Gamma$ denote the space of multi-view association maps, $Z\triangleq(Z^{\left(1:C\right)})$, and assuming that conditional on $\boldsymbol{X}$, these constituent sets are mutually independent, then the \textit{multi-view multi-object measurement likelihood} is given by \cite{vo2019multi}:  
\begin{equation}
\boldsymbol{g}\left(Z|\boldsymbol{X}\right)\propto\sum_{\gamma\in\Gamma}\delta_{\mathcal{L}_{\gamma}}[\mathcal{L}\left(\boldsymbol{X}\right)]\left[\psi_{Z,\boldsymbol{X}}^{(\gamma(\mathcal{L}(\cdot)))}\left(\cdot\right)\right]^{\boldsymbol{X}},\label{e:Multi_Sensor_Lkhd}
\end{equation}
where $\mathcal{L}_{\gamma}\triangleq\{\ell:\gamma^{(1)}(\ell),...,\gamma^{(C)}(\ell)\geq0\}$
denotes the \textit{live label set} of the multi-view association
map $\gamma$, and
\begin{align}
\psi_{Z,\boldsymbol{X}}^{(j^{(1:C)})}\left(\boldsymbol{x}\right) & \triangleq\prod\limits _{c=1}^{C}\psi_{Z^{(c)},\boldsymbol{X}}^{(c,j^{(c)})}\left(\boldsymbol{x}\right).\label{eq:Psi-multiSensor}
\end{align}

\subsection{Bayesian Multi-View MOT Filter}

In Bayesian estimation, the \textit{multi-object filtering density} is the probability density of the current multi-object state conditioned on the observation history. It encapsulates all statistical information on the multi-object state, given the observed data, and prior information described by the multi-object transition density $\boldsymbol{f}(\cdot|\cdot)$ and observation likelihood $\boldsymbol{g}(\cdot|\cdot)$. Multi-object state/trajectory estimate can be determined from the multi-object filtering density via the Joint Multi-object (JoM) or Marginal Multi-object (MaM), including labeled-MaM, estimators \cite{mahler2007statistical}, \cite{vo2023overview}. The latter are commonly used due to their computational tractability. The MaM/labeled-MaM estimate is the most probable (or expected) multi-object state given the most probable cardinality/label-set \cite{mahler2007statistical}, \cite{vo2023overview}.

The multi-object filtering density $\bm{\pi}$ can be propagated forward to the next time via the Bayes recursion 
\begin{equation}
\boldsymbol{\pi}_{+}(\boldsymbol{X}_{+})\thinspace\propto\thinspace\boldsymbol{g}(Z_{+}|\boldsymbol{X}_{+})\int\boldsymbol{f}_{+}(\boldsymbol{X}_{+}|\boldsymbol{X})\boldsymbol{\pi}(\boldsymbol{X})\delta\boldsymbol{X}.\label{eq:bayes_recursion}
\end{equation}
This approach is not only applicable to objects with independent motion, and detection observations, but for more general models including cell mitosis \cite{nguyen2021tracking}, social force model \cite{ishtiaq2023interaction}, track-before-detect \cite{kim2019labeled}, as well as merged measurements \cite{beard2015bayesian}. It also offers the capability to fuse different measurement types, e.g., track-before-detect measurement with detections, simply by multiplying their likelihoods. 

The (exact) Bayes MOT filter (\ref{eq:bayes_recursion}) would fulfill all MOT functionalities. The integration of suitable multi-object dynamic and observation models into the multi-object filtering density allows the filter to initiate/terminate/re-identify tracks, resolve multi-view data association and occlusions, from the observed data. Unfortunately, exact implementation is intractable due mainly to the exponential growth in memory requirement and computational resources. Existing approximations, designed for speed in generic applications, impede MOT functionalities such as track initiation/re-identification and occlusion resolution.

\section{Approximate MV-MOT Filter\label{sec:implement}}

This section presents an approximate Multi-View MOT (MV-MOT) filter that realizes automatic track initiation/re-identification and occlusion resolution by using an adaptive birth model that accounts for reappearing objects and a high-fidelity geometric occlusion model. In Subsection \ref{sec:glmbrecur}, we present a commonly used approximation to the Bayes MV-MOT filter (\ref{eq:bayes_recursion}), which involves a generalized labeled multi-Bernoulli (GLMB) approximation for analytical tractability, and truncating the resulting GLMB components for numerical tractability \cite{ong2020bayesian}. A high-fidelity yet tractable occlusion model based on projections of 3D objects on the camera planes that accommodates full/partial occlusions is developed in Subsection \ref{sec:occlusion}. In Subsection \ref{sec:birthmodel}, we detail an adaptive birth model to realize track initiation and rectify the GLMB truncation to realize re-identification.

\subsection{Multi-View GLMB Recursion\label{sec:glmbrecur}}

This subsection outlines the two-step approximation of the Bayes MV-MOT filter. Consider first the approximation of the multi-object filtering density $\bm{\pi}$, by a GLMB of the form
\begin{equation}
\widehat{\bm{\pi}}\left(\bm{X}\right)=\delta_{|\bm{X}|}\left[\left|\mathcal{L}\left(\bm{X}\right)\right|\right]\sum_{I,\xi}\omega^{\left(I,\xi\right)}\delta_{I}\left[\mathcal{L}\left(\bm{X}\right)\right]\left[p^{\left(\xi\right)}\right]^{\bm{X}},\label{eq:prior_GLMB}
\end{equation}
where: $I$ $\mathcal{\in F}\left(\mathbb{L}\right)$, the \textit{class of all finite subsets of} $\mathbb{L}$; $\xi$ $\in\Xi$, the \textit{space of multi-view association map histories} $\gamma_{1:k}$; each $\omega^{\left(I,\xi\right)}$ is a non-negative weight such that ${\scriptstyle {\textstyle \Sigma}}_{I,\xi}\omega^{\left(I,\xi\right)}=1$; and each $p^{\left(\xi\right)}(\cdot,\ell)$ is a probability density on $\mathbb{X}$. The weight $\omega^{\left(I,\xi\right)}$ can be interpreted as the probability of \textit{hypothesis} $\left(I,\xi\right)$, and conditional on $\left(I,\xi\right)$, $p^{(\xi)}(\cdot,\ell)$ is the probability density of the attribute of $\ell\in I$. A GLMB is completely characterized by its parameters, and hence we adopt the abbreviation
\begin{equation}
\widehat{\bm{\pi}}=\left\{ (\omega^{(I,\xi)},p^{(\xi)}):(I,\xi)\in\mathcal{F}(\mathbb{L})\times\Xi\right\} .\label{eq:GLMB_param}
\end{equation}

Remark 1: Note that the GLMB cardinality distribution, from which we determine the most probable cardinality $n^{*}$ for the MaM estimator, is given by

\[
\textrm{Prob}(|\boldsymbol{X}|=n)=\sum_{I,\xi}\omega^{\left(I,\xi\right)}\delta_{n}[|I|].
\]
For efficiency, instead of computing the most probable multi-object state, we compute the estimated states from $p^{\left(\xi^{*}\right)}(\cdot,\ell)$ for each $\ell\in I^{*}$, where $(I^{*},\xi^{*})$ is the most probable hypothesis such that $|I^{*}|=n^{*}$.

The class of GLMBs is a versatile family of multi-object densities due to its closure under the Bayes recursion (\ref{eq:bayes_recursion}) and efficient approximation (linear complexity in the total number of detections across the sensors) for commonly used multi-object system models \cite{vo2019multi}. Specifically, for $P_{D}^{(1:C)}\left(\boldsymbol{x};\boldsymbol{X}\right)=P_{D}^{(1:C)}\left(\boldsymbol{x}\right)$, if the multi-object filtering density at the current time is a GLMB, then it is also a GLMB at the next time, and given by the MS-GLMB recursion \cite{vo2019multi}
\begin{eqnarray*}
\widehat{\bm{\pi}}_{+} & = & \Omega_{+}(\widehat{\bm{\pi}};P_{D,+}^{(1:C)},\boldsymbol{f}{}_{B,+}),
\end{eqnarray*}
where $\boldsymbol{f}{}_{B,+}\triangleq\{(r_{B,+}^{(\ell)},p_{B,+}^{(\ell)})\}_{\ell\in\mathbb{B}_{+}}$ denotes the parameters of the birth model\footnote{The MS-GLMB recursion also depends on the measurements and other multi-object system parameters, but we suppress them for clarity.}. While the number of GLMB components grows exponentially with time, they can be truncated with minimum $L_{1}$-error, using multi-dimensional rank assignment \cite{vo2014labeled} or Gibbs sampling \cite{vo2017efficient}. Unfortunately, when $P_{D}^{(1:C)}\left(\boldsymbol{x};\boldsymbol{X}\right)\neq P_{D}^{(1:C)}\left(\boldsymbol{x}\right)$ as per our occlusion model, $\bm{\pi}_{+}$ is not a GLMB, and is computationally intractable in general.   

An approximate multi-view GLMB (MV-GLMB) filter for occlusion models with a general $P_{D}^{(1:C)}\left(\boldsymbol{x};\boldsymbol{X}\right)$ has been developed in \cite{ong2020bayesian} by combining piecewise approximation of $P_{D}^{(1:C)}\left(\boldsymbol{x};\boldsymbol{X}\right)$ with importance sampling via the Gibbs sampler. The approximate GLMB filtering density is propagated using the MV-GLMB recursion
\begin{eqnarray}
\widehat{\bm{\pi}}_{+} & = & \widehat{\Omega}(\widehat{\bm{\pi}};P_{D,+}^{(1:C)},\boldsymbol{f}{}_{B,+}),\label{eq:mvglmbrecur}
\end{eqnarray}
summarized in Alg. 2 of \cite{ong2020bayesian}, which extends the MS-GLMB filter to address $P_{D}^{(1:C)}\left(\boldsymbol{x};\boldsymbol{X}\right)\neq P_{D}^{(1:C)}\left(\boldsymbol{x}\right)$.  

Remark 2: The GLMB filtering density can be further approximated by retaining only the best component after each filtering cycle. This approximation uses only the most likely (multi-sensor) measurement-to-track assignment, which is conceptually similar to the strategy of the global nearest-neighbour (GNN) tracker \cite{blackman1999design}. Although this results in considerable improvement in processing speed, performance is expected to degrade, especially in low signal-to-noise scenarios (see also the ablation study in Subsection \ref{subsubsec:best-hypo}).

\begin{figure}[tbh]
\begin{centering}
\includegraphics[width=0.99\columnwidth]{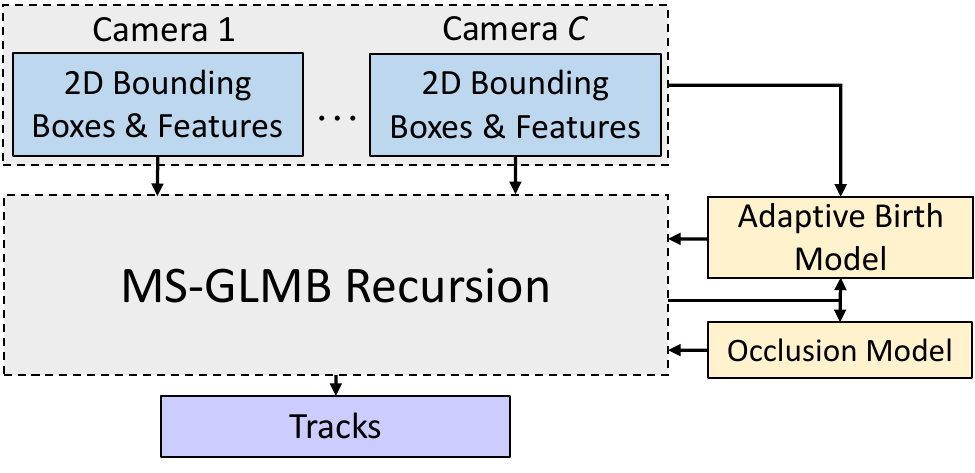}
\par\end{centering}
\caption{Schematic of the proposed multi-view MOT filter, with Adaptive Birth Model and Occlusion Model that realize MOT functionalities\label{fig:mv-glmb-recursion}.}
\end{figure}

Fig. \ref{fig:mv-glmb-recursion} illustrates how our new adaptive birth model and occlusion model is integrated into the MV-GLMB filter to realize the MOT functionalities of (automatic) Track Inititialization, Track Re-Identification, Track Termination, and Occlusion Handling. Details on the proposed occlusion model will be given in the next subsection, and the adaptive estimation of the birth model parameters $\{(r_{B}^{(\ell)},p_{B}^{(\ell)})\}_{\ell\in\mathbb{B}_{+}}$ will be given in Subsection \ref{sec:birthmodel}.

\subsection{Occlusion Modeling\label{sec:occlusion}}

\begin{figure}[tbh]
\begin{centering}
\includegraphics[width=0.99\columnwidth]{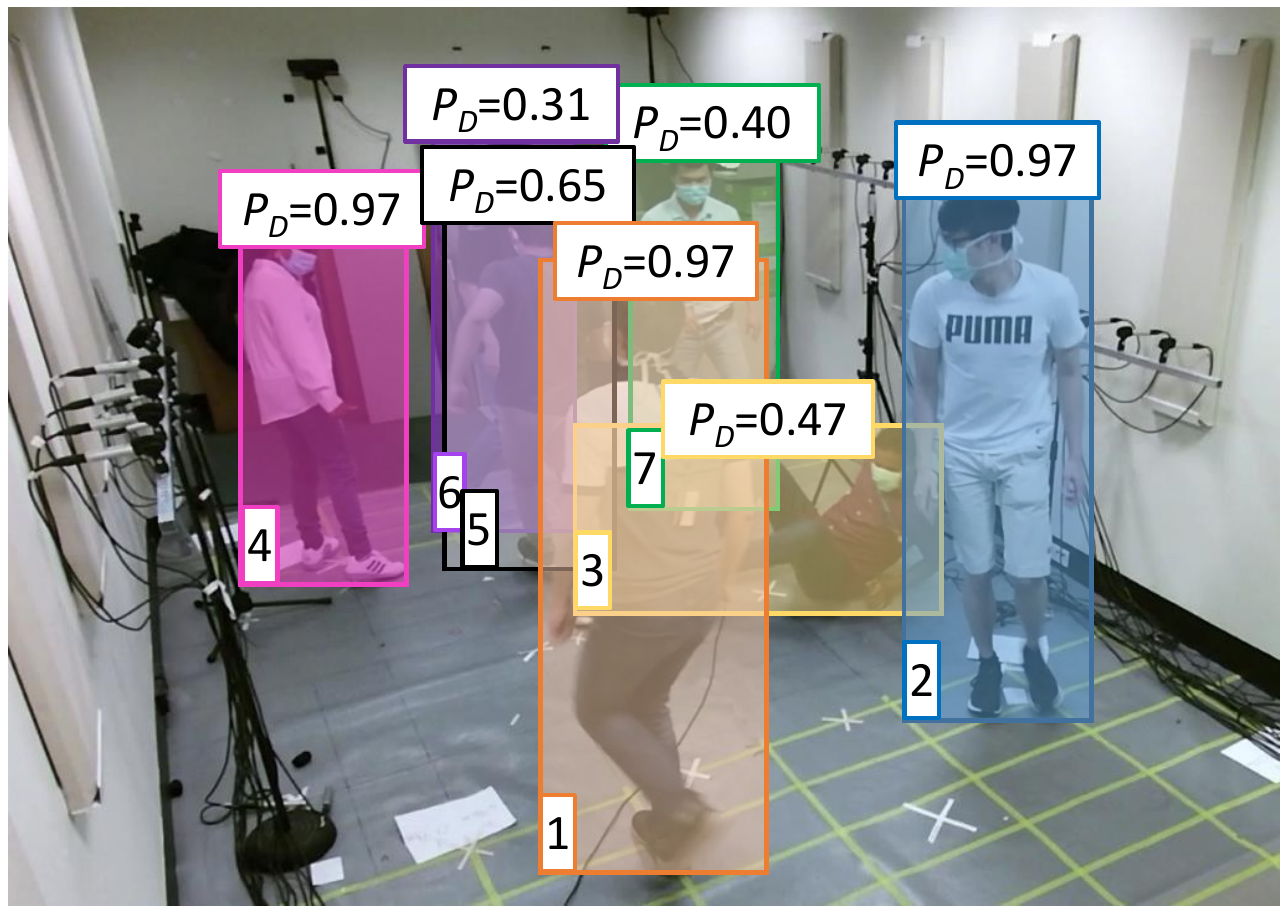}
\par\end{centering}
\caption{For illustration, tracks are indexed from the closest to the furthest from the camera. Track 4 has no overlap with any other tracks and thus has maximum detection probability. Tracks 1 and 2 overlap with other tracks, but closer to the camera (i.e., lower bottom corner), hence they also have maximum detection probability. Track 6 overlaps with track 5, but track 5 has higher detection probability, because it is closer to the camera. \label{fig:ioapd}}
\end{figure}

Rather than using an external occlusion handling module to provide better tracks, the Bayes MOT filter accounts for occlusion via an occlusion model, described by the detection probability of the objects. In the presence of occlusions, the more accurate model is, the better the tracking results. An occlusion model was proposed in \cite{ong2020bayesian}, where the detection probabilities of objects in the shadow regions of others (w.r.t. the LoS of the camera) are assigned small values. While this model is accurate for full occlusions, it does not address partial occlusions, where the detector still has a high probability of detecting the objects.    

In this subsection, we present a new occlusion model that accommodates partial (and full) occlusions. Our model is based on the proportion of area overlap between the boxes bounding the images of the occluded and occluding objects on camera's image plane. The larger the overlap, the lower the detection probability of the occluded object. Noting that an object can only be occluded by those in front of it (i.e., closer to the camera), let $\textrm{Fr}^{(c)}(\boldsymbol{x};\boldsymbol{X})$ denote the subset of objects in $\boldsymbol{X}$ that are in front of $\boldsymbol{x}\in\boldsymbol{X}$ with respect to camera $c$. Then the occlusion score of $\boldsymbol{x}$ is given by 
\begin{equation}
O^{(c)}\left(\boldsymbol{x};\boldsymbol{X}\right)=\frac{\textrm{Area}\left(\Phi^{(c)}(\boldsymbol{x})\bigcap\left(\bigcup_{\boldsymbol{x}'\in\textrm{Fr}^{(c)}(\boldsymbol{x};\boldsymbol{X})}\Phi^{(c)}(\boldsymbol{x}')\right)\right)}{\textrm{Area}\left(\Phi^{(c)}(\boldsymbol{x})\right)},
\end{equation}
where $\Phi^{(c)}(\boldsymbol{x})=\Phi^{(c)}(x)$ for $\boldsymbol{x}=(x,\ell)$, and $\textrm{Area}(S)$ is the area of a 2D shape $S$. Since the more occluded the object is the less likely it will be detected, we use the following the detection probability 
\begin{equation}
P_{D}^{(c)}\left(\boldsymbol{x};\boldsymbol{X}\right)=\max(\epsilon,1-\epsilon-O^{(c)}\left(\boldsymbol{x};\boldsymbol{X}\right)),
\end{equation}
so as to cap it between $\epsilon$ and $1-\epsilon$. Some example detection probability values for a given camera according to this model is shown in Fig. \ref{fig:ioapd}. 

Remark 3: The subset $\textrm{Fr}^{(c)}(\boldsymbol{x};\boldsymbol{X})$ can be determined by comparing the distances of the objects from camera $c$. Alternatively, assuming all objects are on the same ground level, we can compare the lower bottom corners of the bounding boxes of the objects on camera $c$'s plane: those with lower bottom corners are closer to the camera.

\subsection{Adaptive Birth Modeling\label{sec:birthmodel}}

While the MV-GLMB filter can provide automatic track initiation and re-identification, it requires a combination of accurate prior birth model (that varies with time) and prudent approximation. In this subsection, we develop a tractable technique to estimate birth model online and rectify the GLMB truncation process to realize track initiation and re-identification.

\subsubsection{Adaptive Birth Model Parameters \label{sec:adaptbirth}}

In \cite{trezza2022multi}, an efficient technique was developed for estimating the LMB birth model parameters $\{(r_{B,+}^{(\ell_{+})},$ $p_{B,+}^{(\ell_{+})})\}_{\ell\in\mathbb{B}_{+}}$ (Section \ref{subsec:MOT-transition}), using the current multi-sensor measurement. This approach seeks an empirical LMB birth model that provides a good fit of the multi-camera measurement $Z$. Given the current GLMB filtering density (\ref{eq:GLMB_param}), suppose that the multi-camera measurement $Z$ is generated from new birth objects according to the multi-view association map $\mathring{\gamma}:\mathbb{B}_{+}\rightarrow\{-1\}^{C}\uplus(\mathbb{J}^{(1)}\times\cdots\times\mathbb{J}^{(C)})$. Then, the best fitting empirical LMB model is given by \cite{trezza2022multi}
\begin{equation}
\{(\widehat{r}_{\mathring{\gamma}}^{(\ell_{+})},\widehat{p}_{\mathring{\gamma}}^{(\ell_{+})})\}_{\ell_{+}\in\mathcal{L}_{\mathring{\gamma}}},\label{eq:empirical_LMB_birth}
\end{equation}
where: $\mathcal{L}_{\mathring{\gamma}}$ is the live label set of $\mathring{\gamma}$;
\begin{eqnarray*}
\widehat{r}_{\mathring{\gamma}}^{(\ell_{+})} & \!\!\!= & \!\!\!\min\!\left(\!r_{B,+}^{*},\frac{\lambda_{B,+}r_{U\!}\left(\mathring{\gamma}(\ell_{+})\right)\!\bar{\psi}_{Z,B}^{\left(\mathring{\gamma}(\ell_{+})\right)\!}\left(\ell_{+}\right)}{\langle\!r_{U\!}\left(\mathring{\gamma}(\cdot)\right),\bar{\psi}_{Z,B}^{\left(\mathring{\gamma}(\cdot)\right)\!}\left(\cdot\right)\rangle}\!\right);\\
\widehat{p}_{\mathring{\gamma}}^{(\ell_{+})}(x_{+}) & \!\!\!\propto & \!\!\!\int\!f_{+}(x_{+}|x,\ell_{+})p_{B,0}^{(\ell_{+})}(x)\psi_{Z,B}^{\left(\mathring{\gamma}(\ell_{+})\right)}(x,\ell_{+})dx;
\end{eqnarray*}
$r_{B,+}^{*}$ is a prescribed maximum birth probability; $\lambda_{B,+}$ is a prescribed expected number of births;
\begin{eqnarray}
r_{U}(j^{(1:C)}) & = & \prod_{c=1}^{C}\left[1-\sum_{I,\xi}1_{\xi^{(c)}(I)}(j^{(c)})\omega^{\left(I,\xi\right)}\right];\\
\bar{\psi}_{Z,B}^{(j^{(1:C)})}(\ell) & = & \langle p_{B,0}(\cdot,\ell),\psi_{Z,B}^{(j^{(1:C)})}(\cdot,\ell)\rangle;
\end{eqnarray}
$\xi^{(c)}(I)=\{\gamma^{(c)}(\ell):\ell\in I\}$; $p_{B,0}(x,\ell)$ is a prescribed prior birth probability density; $\psi_{Z,B}^{(j^{(1:C)})\!}\left(x,\ell\right)$ is $\psi_{Z,\boldsymbol{X}}^{(j^{(1:C)})\!}\left(x,\ell\right)$ in (\ref{e:Multi_Sensor_Lkhd}) with $P_{D}^{\left(c\right)}\left(\boldsymbol{x};\boldsymbol{X}\right)$ set to a prescribed (constant) detection probability $P_{D,B}^{\left(c\right)}$, and the feature likelihood $g_{f}^{(c)}(z_{f}^{(c)}|\ell)$ in the single-view single-object measurement likelihood (\ref{eq:solikelihood-whole}) set to a uniform distribution (hence, only the bounding box measurements are used in the birth model estimation). 

The empirical LMB birth (\ref{eq:empirical_LMB_birth}) is completely parameterized by the multi-view association map $\mathring{\gamma}$. This birth model reduces prior knowledge on a large number of LMB model parameters to only four prescribed parameters $r_{B,+}^{*}$, $\lambda_{B,+}$ (usually set to 1), $p_{B,0}(x,\ell)$ and $P_{D,B}$. Intuitively, the componenst of a good fitting empirical LMB should have significant existence probabilities. 

Remark 4: Note that in this work, we used the multi-view association map $\mathring{\gamma}=(\mathring{\gamma}^{\left(1:C\right)})$ instead of the injection $\theta_{B}:\mathbb{J}^{(1)}\times\cdots\times\mathbb{J}^{(C)}\rightarrow\mathbb{B}_{+}$ in \cite{trezza2022multi}. The key difference is that $\mathring{\gamma}$ constrains each single-camera detection to originate from at most one object (see Subsection \ref{subsec:singlemeas}). In contrast, $\theta_{B}$ relaxes this constraint and allows each single-camera detection to originate from multiple objects, which can result in increased false track initiations, especially for scenarios with large areas. Nonetheless, this relaxation enables the authors to develop a Gibbs sampler to compute a good fitting (empirical LMB parameterized by) $\theta_{B}$ \cite{trezza2022multi}.  

Since the Gibbs sampler in \cite{trezza2022multi} cannot accommodate the constraint of at most one object per detection, we use clustering to determine a good fitting (empirical LMB parameterized by) $\mathring{\gamma}$. Intuitively, the detections generated by the same object at every camera would be clustered around the object's position when projected into the ground plane. Hence, the $\mathring{\gamma}$ constructed by clustering (single-camera) detections in the ground plane so that each cluster corresponds to detections generated by an object, provides a good fit to the multi-camera measurement $Z$. Note that since $\mathring{\gamma}$ is multi-camera association map, each detection can only belong to at most one cluster.    

The clustering algorithm is described in Alg. \ref{alg:Mean-shift-clustering}. The multi-view association map $\mathring{\gamma}$, represented as an assignment matrix where each row consists of the measurement indices that belong to one cluster. In step one, a set of initial cluster means is generated in a similar manner to the popular mean shift clustering algorithm. In step two, $\mathring{\gamma}$ is constructed by sequentially appending each row of the associated measurement indices. In the pseudocode, the 'TransformToGroundPlane' function is a homography transformation taking 2D measurements to their ground plane positions. The 'dist' function computes the distance between points in the ground plane. The 'ComputeCentroid' function returns the centroid in the ground plane with inputs as a list of points and the corresponding indices specifying which points are used to compute the centroid.

\subsubsection{Track Initialization and Re-Identification\label{sec:trackinitreid}}

The birth model enables the MV-GLMB recursion (\ref{eq:mvglmbrecur}) to automatically initiates new tracks, and in principle, re-identify reappearing tracks. Labels that have ever existed (up to the current time) are captured in some components of the (untruncated) GLMB filtering density. When new data arrives, the MV-GLMB recursion updates their existence probabilities accordingly so that those in the scene have high existence probabilities and vice-versa. However, in practice, component truncation deletes labels with prolonged low existence probabilities permanently from the GLMB density. This means they cannot be recovered even when new data support their reappearance, and each LMB birth parameter in $\{(\widehat{r}_{\mathring{\gamma}}^{(\ell)},\widehat{p}_{\mathring{\gamma}}^{(\ell)})\}_{\ell\in\mathcal{L}_{\mathring{\gamma}}}$ could either correspond to a new track, or reappearing track.  

To restore the filter's track re-identification functionality, we propose to retain tracks that would have been deleted in the the GLMB truncation, herein referred to as \textit{Tentatively Terminated} (TT) tracks, and relabel the subset of $\{(\widehat{r}_{\mathring{\gamma}}^{(\ell)},\widehat{p}_{\mathring{\gamma}}^{(\ell)})\}_{\ell\in\mathcal{L}_{\mathring{\gamma}}}$ with the labels of the TT tracks that best match them in visual features\footnote{Visual features are better suited for re-identification because they are relatively stable over time \cite{zhang2021fairmot}, whereas kinematic and shapes attributes vary with time, while tracks reappear almost independently from the locations where they disappeared. }. A TT track retains the visual feature from its corresponding label in highest weighted GLMB component at the time of TT, and will only be permanently deleted if it is not re-identified within a prescribed period. Note that after relabeling we remove the corresponding TT track from the TT set. The GLMB recursion with the relabeled LMB birth model will update the existence probabilities of the reappearing and new birth tracks in accordance with the received multi-camera data. 

Optimal assignment can be used to match the live labels $\mathcal{L}_{\mathring{\gamma}}$ and the TT tracks, and only those with matching scores above a recall threshold $\tau_{R}$ are relabeled. The matching score of a label $\ell_{+}\in\mathcal{L}_{\mathring{\gamma}}$ to a TT label $\ell=(s,\iota)$ with feature $\alpha^{(\ell,c)}$ from each camera $c\in\{1:C\}$, is defined as
\begin{equation}
R_{\ell_{+},\ell}=\frac{k-s}{e(\ell)}\max_{c\in\{1:C\}}s_{f}\left(f(z_{^{\mathring{\gamma}^{(c)}(\ell_{+})}}^{(c)}),\alpha^{(\ell,c)}\right),\label{eq:recallcost}
\end{equation}
where $e(\ell)$ denotes the number of times that label $\ell$ is included in the GLMB density but not as a TT track, $f(z_{j}^{(c)})$ denotes the feature component of the $j$-th measurement from camera $c$, and $s_{f}(\cdot,\cdot)$ is the similarity measure between two feature vectors, see also (\ref{eq:mvglmbrecur}). The rationale behind the time ratio in (\ref{eq:recallcost}) is that the longer the label exists in the GLMB density, the more likely it still exists even though it is TT.

\begin{algorithm}[tbh]
\setstretch{0.2}

\textbf{Input}: $Z$, $h_{1:C}$, $\epsilon$ \\

\textbf{Output}: $\mathring{\gamma}$ 

\rule[0.5ex]{0.95\columnwidth}{0.5pt} \\
 \textit{Step 1: Generate Initial Cluster Means}\\
$S=\emptyset$ 

\For{$c=1:C$}{ $S^{(c)}=\text{TransformToGroundPlane}(Z^{(c)})$
\\
 $S=S\cup S^{(c)}$\\
 } 
\While{all elements in  $S$ \textbf{not} converge}{ \For{ $g=1:|S|$}{
\If{$S[g]$ \textbf{not} converge}{ $P=\emptyset$;\\ 
 \For{ $c=1:C$}{
 \For{ $i=1:|S^{(c)}|$}{ \If{$\text{dist}(S^{(c)}[i],S[g])<h_{c}$}{
$P=P\cup S^{(c)}[i]$ } } } $new\_S_{g}\gets K\left(\text{min}(P)-S[g]\right)\text{min}(P)$ \\
\If{dist($(S[g],new\_S_g)<\epsilon$}{$S[g]$ is converge} 
$S[g]\gets new\_S_{g}$} } } 

\textit{Step 2: Generate Multi-View Association Map}\\
$\mathring{\gamma}=\emptyset$; $t=0$;

\For{$c=1:C$}{ \For{$i=1:|S^{(c)}|$}{ $l=-1$; $p=+\infty$;
$t=t+1$;\\
 \For{$j=1:\text{NumberOfRow}(\mathring{\gamma})$}{ 
 $cen$ = ComputeCentroid($\mathring{\gamma}[j,:]$, $S^{(1:C)}$)\\
 $d$ = dist$\left(S^{(c)}_{i},cen\right)$;\\
 \If{$(d\!<\!h_{c})$ $\!\!\land\!\!$ $(d\!<\!p)$
$\!\!\land\!\!$ $(\mathring{\gamma}[j,c]\!=\!0)$}{ $l=j$; $p=d$;
} } \eIf{$l=-1$}{ $M=0_{1\times C}$;
$M_{c}=i$;\\
 $\mathring{\gamma}=\text{AppendRow}(\mathring{\gamma}, M)$; 
 }{ $\mathring{\gamma}[l,i]=i$; 
 } } }

\caption{Clustering for Adaptive Birth. \label{alg:Mean-shift-clustering}}
\end{algorithm}

\begin{figure*}[t]
    % \vspace{-0.6cm}
    \begin{centering}
        % \raisebox{0.5in}{\rotatebox[origin=t]{90}{MV-GLMB-MS}}
        \includegraphics[width=\textwidth]{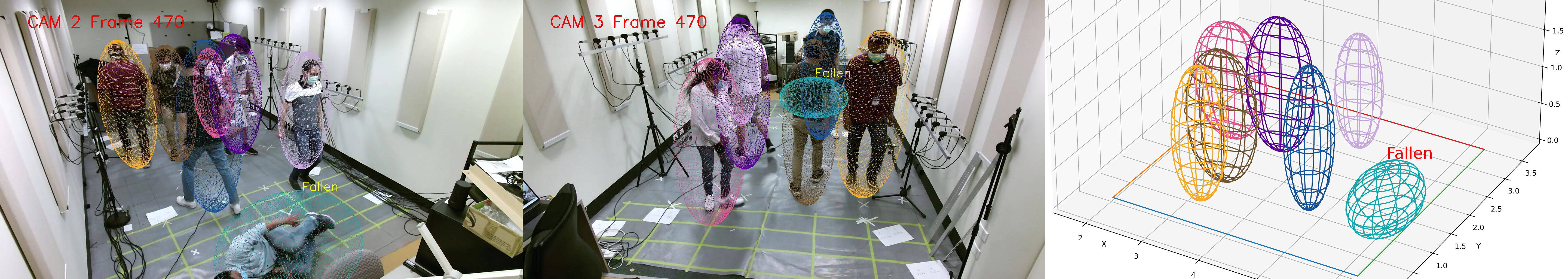}
        % \raisebox{0.5in}{\rotatebox[origin=t]{270}{CMC5}}
    % \end{centering}
    \vskip 0.05cm
    % \begin{centering}
        % \raisebox{0.5in}{\rotatebox[origin=t]{90}{MV-GLMB-MS}}
        \includegraphics[width=\textwidth]{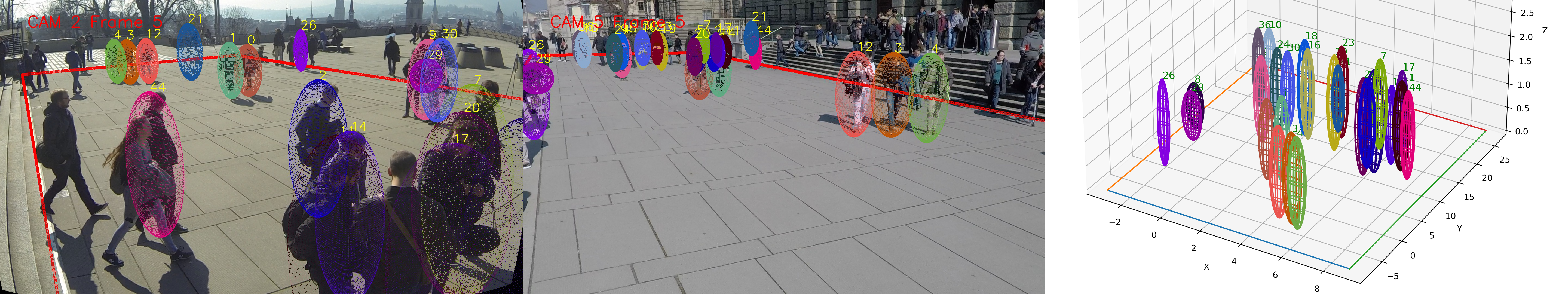}
        % \raisebox{0.5in}{\rotatebox[origin=t]{270}{Wildtrack}}
    \end{centering}
    
    \caption{3D ellipsoid estimates from the proposed MV-GLMB-AB filter using CSTrack detection inputs. Top row: CMC5 dataset at frame 470 for cameras 2 and 3. Bottom row: WT dataset at frame 25 for cameras 2 and 5 (only objects inside the red boundary are considered). The first two columns show the projected 3D estimates on the respective camera planes, and the last column shows the 3D estimates. Each color corresponds to a unique object ID. Videos are also provided in the supplementary materials.\label{fig:quali}}
\end{figure*}

\begin{figure*}[ht]
\includegraphics[width=1\textwidth]{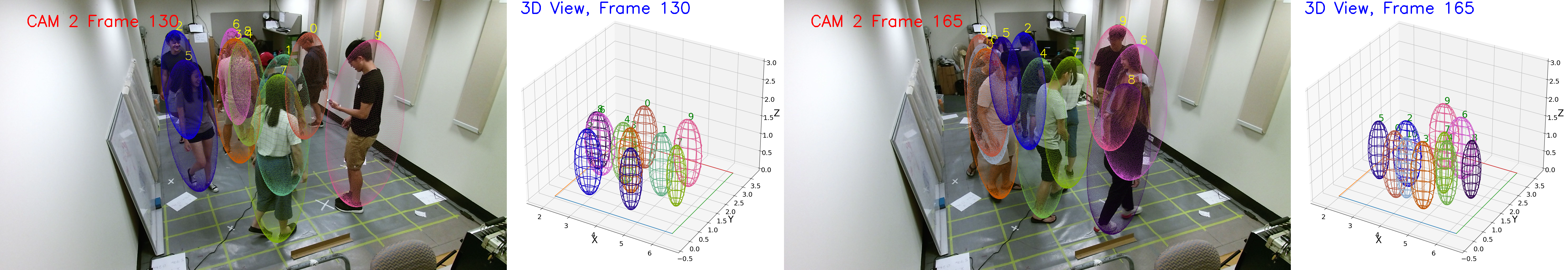}
\vskip .05cm 
\includegraphics[width=1\textwidth]{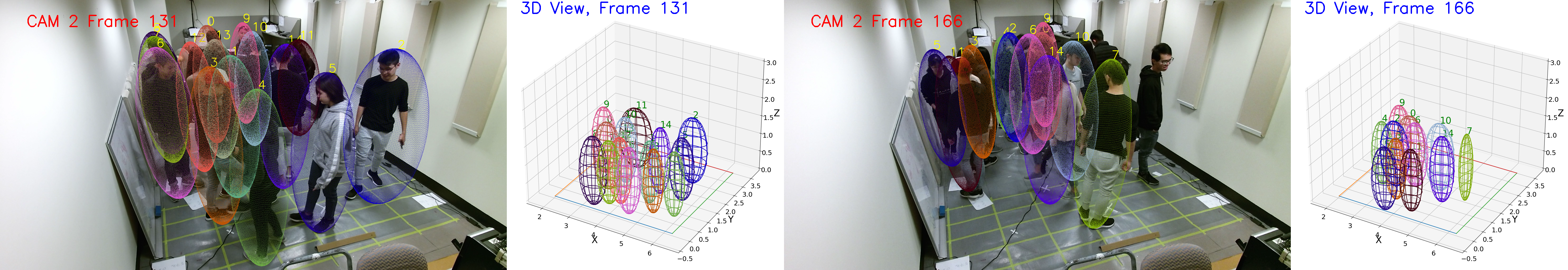}
\vskip .05cm
\includegraphics[width=1\textwidth]{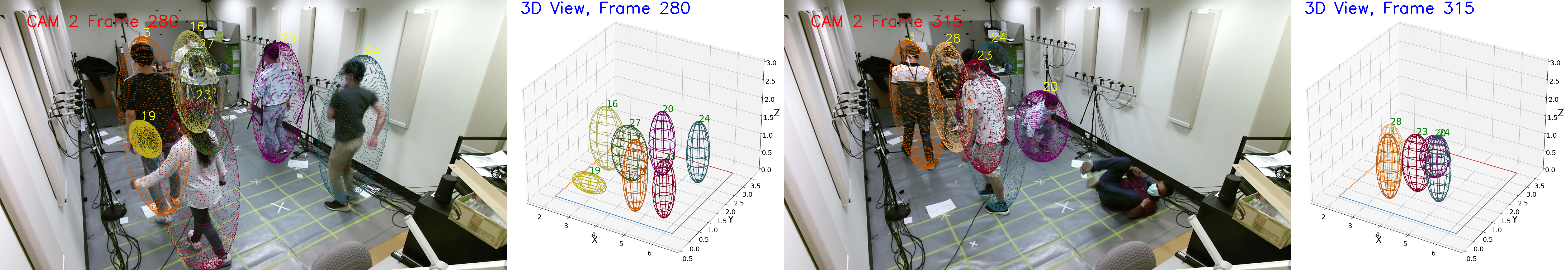}
\caption{Track re-identification: 3D ellipsoid estimates from the MV-GLMB-AB filter using CSTrack detection. Object disappearance-reappearance (in CMC) is simulated by turning off all cameras mid-scene for 30 frames. Top row: CMC2-all cameras off from frames 130-160. Middle row: CMC3-all cameras off from frames 131-161. Bottom row: CMC5-all cameras off from frames 280-310. Columns 1 and 2 show estimates, projected on camera 2 and in 3D, before turning off all cameras. Columns 3 and 4 show the estimates 5 frames after all cameras are turned back on.\label{fig:quali_cmcoff30f}}
\end{figure*}

\section{Experimental Results}\label{sec:experiment}

This section presents experimental evaluations of our proposed 3D MV-MOT algorithm, referred to as Multiview GLMB-Adaptive Birth (MV-GLMB-AB) filter, on the WILDTRACK (WT) \cite{chavdarova2018wildtrack} dataset (with 7 cameras) and the Curtin multi-camera (CMC) dataset (with 4 cameras, and five sequences CMC1-5) \cite{ong2020bayesian}. While the WT dataset is popular for 3D MOT, it only provides the \textit{true ground plane} positions of the objects, \textit{not their 3D positions} and extents. The CMC dataset fills this gap, with details on height, width, and 3D position. Sequences 1-5 of the CMC dataset have different object densities for performance evaluation with varying levels of difficulties. Further, sequences 4 and 5 also include jumping and falling people to evaluate pose estimation capability.  

To quantify tracking performance, we use the track identity measure \cite{ristani2016performance}, including IDF1 score, numbers of mostly tracked (MT) tracks, partially tracked (PT) tracks, and ID switches (IDS). We also report the CLEAR MOT measure \cite{bernardin2008evaluating}, including multi-object tracking accuracy (MOTA) score, numbers of false positive (FP), false negative (FN), and OSPA\!$^{\texttt{(2)}}$\! error \cite{beard2020asolution,rezatofighi2020trustworthy}. For evaluations that only considered ground plane positions of the tracks, the Euclidean distance is used as the base-distance. For evaluations of 3D trajectories with extents (a 3D ellipsoid), the GIoU distance (normalized to the interval {[}0,1{]}) between 3D bounding boxes is used as the base-distance. The evaluation threshold is 1 meter for the Euclidean distance and 0.5 for the GIoU distance. The cut-off distance for the OSPA\!$^{\texttt{(2)}}$\! metric is set to 1 meter for Euclidean base-distance and to 1 for GIoU base-distance. Since the filters involve random sampling, we evaluate their performance over 25 Monte Carlo (MC) runs and report the mean and standard deviation of the results. 

In subsection \ref{sec:perforeval} a comparison of tracking performance and run-time between the MV-GLMB-AB filter and other state-of-the-art methods, demonstrating the ability to re-identify tracks and uninterrupted operations when cameras are added, removed, or repositioned on-the-fly. Subsection \ref{sec:benchmarkideal} benchmarks the tracking performance of MV-GLMB-AB filter against baseline single-sensor filters that process ground plane measurements from ideal detectors, trained on the ground truth dataset. An ablation study on the models used in the MV-GLMB-AB filter is presented in Subsection \ref{subsec:exp-ablation}.

The following model parameters are used for all experiments in this section. The dynamic noise variance (measured in squared meters) is set to $\upsilon^{(\zeta)}$=$[0.0012,0.0012,0.0012]^{T}$ for the CMC dataset and $\upsilon^{(\zeta)}$=$[0.0225,0.0225,0.0225]^{T}$ for the WT dataset, while $\upsilon^{(\varsigma)}$=$[0.0036,0.0036,0.0004]^{T}$ for both. For each camera, the measurement noise variance (measured in pixels) is set to $\upsilon_{p}^{(c)}=[400,400]^{T}$, and $\upsilon_{e}^{(c)}$=$[0.00995,$ $0.0025]^{T}$ for upright objects and $\upsilon_{e}^{(c)}$=$[0.0025,0.00995]^{T}$ for fallen objects. Since we observe a low number of false positive measurements in the tested scenarios, we set the clutter rate to 5 for all sensors in our implementation. If a higher number of false positive measurements is observed, the clutter rate can be estimated on-the-fly from the data.

\subsection{Performance Evaluation\label{sec:perforeval}}

\subsubsection{Fixed multi-camera configuration}

We compare the proposed filter with current 3D MOT algorithms that process 2D multi-view detections, namely the MV-GLMB \cite{ong2020bayesian} and MS-GLMB \cite{vo2019multi} filters. Note that the CMC4-5 sequences are used to compare the performance on scenarios involving human poses (upright or fallen) with the MV-GLMB filter. 

The proposed filter's output on typical scenarios in the CMC5 sequence and the WT dataset with CSTrack detections are shown in Fig.~\ref{fig:quali}. For the CMC5 sequence, the proposed MOT filter yields accurate object positions and poses. The relatively poor detection quality of the CSTrack detector (see Tab. \ref{tbl:detquality-WT}) on the WT dataset is manifested in several misdetections in the proposed filter's output. 

Fig.~\ref{fig:quali_cmcoff30f} illustrates the filter's re-identification capability, where object disappearance-reappearance is simulated by turning off all cameras mid-scene for 30 frames so that most or all of the tracks are terminated before the cameras are on again. Lost 3D tracks are re-identified after they reappear using feature information from monocular camera images. Recall performance is best when the features are stable and unique across frames as per CMC1 and CMC2 (see also video supplementary material). Recall is poor when the objects have similar appearance (i.e., non-unique features) as per CMC3, or change their appearance quickly (i.e., unstable features) as per CMC5 with pose changes.

%\textcolor{magenta}{Similar behaviour on CMC4 is illustrated in the videos of the supplementary material.}\textcolor{blue}{{} }\textcolor{red}{-may need to move some of these observation to the quantitative as appropriate}\textcolor{blue}{. }\textcolor{red}{Similar behaviour on other CMC sequences are illustrated in the videos of the supplementary material-please update this because I don't know which vidoes you submit.}

\begin{table*}[tbh]
\centering
\global\long\def\arraystretch{1.3}%
 \caption{Tracking performance (in the \textit{ground plane}) on the WT dataset with the CSTrack detector: MC means and 1 standard deviation (shown in parenthesis, only reported for the main measures). The best result for each sequence is \textbf{Bolded}. \label{tbl:wildtrack}}
 \scriptsize
\begin{tabular}{|c|c|ccccccccc|}
\hline 
Detector & Tracker & MT$\uparrow$ & PT$\downarrow$ & ML$\downarrow$ & FP$\downarrow$ & FN$\downarrow$ & IDs$\downarrow$ & MOTA$\uparrow$ & IDF1$\uparrow$ & OSPA\!$^{\texttt{(2)}}$\!$\downarrow$\tabularnewline
\hline 
\hline 
\multirow{3}{*}{CSTrack } & Ours & \textbf{101} & \textbf{134} & \textbf{77} & 5040 & \textbf{2804} & \textbf{273} & \textbf{14.7(3.37)} & \textbf{50.9(2.11)} & \textbf{0.79(0.01)}\tabularnewline
 & MV-GLMB  & 57 & 152 & 103 & 3916 & 4161 & 422 & 10.7(2.13) & 33.2(1.07) & 0.86(0.01)\tabularnewline
 & MS-GLMB  & 58 & 153 & 100 & \textbf{3890} & 4148 & 419 & 11.1(2.06) & 33.3(0.97) & 0.86(0.01)\tabularnewline
\hline 
\multirow{3}{*}{FairMOT } & Ours & \textbf{119} & \textbf{127} & \textbf{66} & 3399 & \textbf{2463} & \textbf{215} & \textbf{36.1(2.60)} & \textbf{58.4(1.98)} & \textbf{0.73(0.01)}\tabularnewline
 & MV-GLMB  & 45 & 147 & 120 & \textbf{3237} & 4353 & 383 & 16.2(1.79) & 31.8(0.77) & 0.86(0.00)\tabularnewline
 & MS-GLMB  & 42 & 146 & 123 & 3260 & 4404 & 379 & 15.5(1.48) & 31.4(0.8) & 0.87(0.00)\tabularnewline
\hline 
\end{tabular}
\end{table*}

\begin{table*}[tbh]
\centering
\global\long\def\arraystretch{1.3}%
 \caption{Tracking performance (in \textit{3D and ground plane}) on the CMC dataset with the CSTrack detector: MC means and 1 standard deviation (shown in parenthesis, only reported for the main measures). The best result for each sequence is \textbf{Bolded}. \label{tbl:cmc}}
 \scriptsize
\begin{tabular}{|>{\centering}p{0.8cm}|c|P{0.2cm}P{0.2cm}P{0.2cm}P{1.05cm}P{1.05cm}P{1.2cm}|P{0.2cm}P{0.2cm}P{0.2cm}P{1.05cm}P{1.05cm}P{1.2cm}|}
\hline 
\multirow{2}{*}{\centering{}Seq.} & \multirow{2}{*}{Tracker} & \multicolumn{6}{c|}{Evaluation with 3D ellipsoid estimates} & \multicolumn{6}{c|}{Evaluation with ground plane estimates}\tabularnewline
\cline{3-14} \cline{4-14} \cline{5-14} \cline{6-14} \cline{7-14} \cline{8-14} \cline{9-14} \cline{10-14} \cline{11-14} \cline{12-14} \cline{13-14} \cline{14-14} 
 &  & FP$\downarrow$ & FN$\downarrow$ & IDs$\downarrow$ & MOTA$\uparrow$ & IDF1$\uparrow$ & OSPA\!$^{\texttt{(2)}}$\!$\downarrow$ & FP$\downarrow$ & FN$\downarrow$ & IDs$\downarrow$ & MOTA$\uparrow$ & IDF1$\uparrow$ & OSPA\!$^{\texttt{(2)}}$\!$\downarrow$\tabularnewline
\hline 
\hline 
\multirow{3}{*}{CMC1} & Ours & \textbf{0} & 4 & \textbf{0} & \textbf{99.4(0.00)} & \textbf{99.7(0.00)} & \textbf{0.3(0.00)} & \textbf{0} & 4 & \textbf{0} & \textbf{99.4(0.00)} & \textbf{99.7(0.00)} & \textbf{0.07(0.00)}\tabularnewline
 & MV-GLMB  & 47 & 3 & \textbf{0} & 92.1(2.86) & 96.0(1.49) & 0.83(0.02) & 47 & \textbf{2} & 1 & 92.2(2.89) & 96.0(1.50) & 0.78(0.02)\tabularnewline
 & MS-GLMB  & 20 & \textbf{2} & \textbf{0} & 96.4(1.97) & 98.1(1.11) & 0.82(0.01) & 20 & \textbf{2} & \textbf{0} & 96.5(1.98) & 98.1(1.11) & 0.76(0.02)\tabularnewline
\hline 
\multirow{3}{*}{CMC2} & Ours & \textbf{18} & \textbf{36} & \textbf{5} & \textbf{97.1(1.43)} & \textbf{90.2(6.62)} & \textbf{0.41(0.03)} & \textbf{16} & 33 & \textbf{4} & \textbf{97.4(1.38)} & \textbf{94.0(4.28)} & \textbf{0.26(0.05)}\tabularnewline
 & MV-GLMB  & 353 & 41 & 56 & 78.3(2.76) & 52.4(6.79) & 0.88(0.02) & 337 & \textbf{26} & 46 & 80.2(2.86) & 64.4(5.92) & 0.87(0.02)\tabularnewline
 & MS-GLMB  & 141 & 126 & 78 & 83.3(2.71) & 47.8(5.06) & 0.88(0.02) & 105 & 89 & 57 & 87.8(1.99) & 60.9(4.71) & 0.87(0.02)\tabularnewline
\hline 
\multirow{3}{*}{CMC3} & Ours & \textbf{57} & \textbf{99} & \textbf{15} & \textbf{93.9(1.15)} & \textbf{78.6(5.03)} & \textbf{0.45(0.03)} & \textbf{28} & 70 & \textbf{14} & \textbf{96.0(1.02)} & \textbf{85.8(3.63)} & \textbf{0.33(0.04)}\tabularnewline
 & MV-GLMB  & 572 & 137 & 105 & 71.2(3.61) & 43.8(3.76) & 0.86(0.02) & 489 & \textbf{55} & 87 & 77.6(2.88) & 58.0(3.18) & 0.85(0.02)\tabularnewline
 & MS-GLMB  & 315 & 327 & 142 & 72.2(4.94) & 38.7(2.59) & 0.89(0.01) & 181 & 192 & 110 & 82.9(4.63) & 55.6(2.65) & 0.89(0.01)\tabularnewline
\hline 
\multirow{3}{*}{CMC4} & Ours & \textbf{0} & \textbf{9} & \textbf{0} & \textbf{97.5(0.29)} & \textbf{98.7(0.16)} & \textbf{0.24(0.00)} & \textbf{0} & \textbf{9} & \textbf{0} & \textbf{97.8(0.00)} & \textbf{98.9(0.00)} & \textbf{0.11(0.00)}\tabularnewline
 & MV-GLMB  & 19 & 94 & 3 & 70.9(4.00) & 66.6(3.31) & 0.71(0.05) & 8 & 83 & 4 & 76.0(3.30) & 68.8(3.16) & 0.67(0.06)\tabularnewline
 & MS-GLMB  & 70 & 103 & 3 & 56.2(15.93) & 74.6(6.90) & 0.66(0.06) & 60 & 92 & 4 & 61.1(15.79) & 76.9(6.77) & 0.62(0.07)\tabularnewline
\hline 
\multirow{3}{*}{CMC5} & Ours & \textbf{83} & \textbf{328} & \textbf{27} & \textbf{88.1(0.69)} & \textbf{50.9(3.02)} & \textbf{0.87(0.02)} & \textbf{32} & \textbf{277} & \textbf{24} & \textbf{91.0(0.51)} & \textbf{51.9(3.04)} & \textbf{0.86(0.02)}\tabularnewline
 & MV-GLMB  & 601 & 597 & 102 & 65.0(7.06) & 24.7(2.47) & 0.93(0.01) & 411 & 408 & 95 & 75.4(6.73) & 31.7(2.60) & 0.94(0.01)\tabularnewline
 & MS-GLMB  & 456 & 690 & 155 & 65.0(6.53) & 20.3(3.32) & 0.96(0.01) & 210 & 444 & 144 & 78.5(5.47) & 27.7(4.14) & 0.97(0.01)\tabularnewline
\hline 
\end{tabular}
\end{table*}

\begin{table*}[tbh]
\centering
% \global\long\def\arraystretch{1.3}%
 \caption{Tracking performance (in \textit{3D and ground plane}) on the CMC dataset with disappearing-reappearing objects, and CSTrack detector: MC means and 1 standard deviation (shown in parenthesis, only reported for the main measures). Object disappearance-reappearance is simulated by turning off all cameras mid-scene for 30 frames. The best result for each sequence is \textbf{Bolded}\label{tbl:cmc-turnof30f}}
 \scriptsize
\begin{tabular}{|>{\centering}p{0.8cm}|c|P{0.2cm}P{0.2cm}P{0.2cm}P{1.1cm}P{1.05cm}P{1.2cm}|P{0.2cm}P{0.2cm}P{0.2cm}P{1.1cm}P{1.05cm}P{1.2cm}|}
\hline 
\multirow{2}{0.6cm}{\centering{}Seq.} & \multirow{2}{*}{Tracker} & \multicolumn{6}{c|}{Evaluation with 3D ellipsoid estimates} & \multicolumn{6}{c|}{Evaluation with ground plane estimates}\tabularnewline
\cline{3-14} \cline{4-14} \cline{5-14} \cline{6-14} \cline{7-14} \cline{8-14} \cline{9-14} \cline{10-14} \cline{11-14} \cline{12-14} \cline{13-14} \cline{14-14} 
 &  & FP$\downarrow$ & FN$\downarrow$ & IDs$\downarrow$ & MOTA$\uparrow$ & IDF1$\uparrow$ & OSPA\!$^{\texttt{(2)}}$\!$\downarrow$ & FP$\downarrow$ & FN$\downarrow$ & IDs$\downarrow$ & MOTA$\uparrow$ & IDF1$\uparrow$ & OSPA\!$^{\texttt{(2)}}$\!$\downarrow$\tabularnewline
\hline 
\hline 
\multirow{3}{*}{CMC1} & Ours & \textbf{0} & \textbf{85} & \textbf{0} & \textbf{87.0(0.14)} & \textbf{93.1(0.10)} & \textbf{0.38(0.00)} & \textbf{0} & \textbf{85} & \textbf{0} & \textbf{87.0(0.17)} & \textbf{93.1(0.10)} & \textbf{0.19(0.00)}\tabularnewline
 & MV-GLMB & 44 & 92 & 3 & 78.6(2.56) & 51.2(0.93) & 0.93(0.01) & 41 & 89 & 4 & 79.5(2.60) & 51.2(0.91) & 0.91(0.01)\tabularnewline
 & MS-GLMB & 25 & 91 & 4 & 81.5(2.08) & 51.9(0.69) & 0.93(0.00) & 23 & 89 & 4 & 82.0(2.00) & 51.9(0.68) & 0.91(0.01)\tabularnewline
\hline 
\multirow{3}{*}{CMC2} & Ours & \textbf{48} & \textbf{290} & \textbf{18} & \textbf{82.8(0.58)} & \textbf{74.0(6.01)} & \textbf{0.54(0.04)} & \textbf{13} & \textbf{254} & \textbf{16} & \textbf{86.3(0.7)} & \textbf{80.1(4.37)} & \textbf{0.45(0.05)}\tabularnewline
 & MV-GLMB & 334 & 360 & 75 & 63.0(3.12) & 36.5(2.43) & 0.94(0.01) & 279 & 306 & 64 & 68.7(2.79) & 43.3(2.09) & 0.93(0.01)\tabularnewline
 & MS-GLMB & 170 & 445 & 105 & 65.3(2.28) & 32.8(1.90) & 0.95(0.00) & 102 & 377 & 86 & 72.7(1.13) & 41.0(1.87) & 0.95(0.00)\tabularnewline
\hline 
\multirow{3}{*}{CMC3} & Ours & \textbf{110} & \textbf{446} & \textbf{39} & \textbf{78.9(1.05)} & \textbf{55.4(2.83)} & \textbf{0.69(0.02)} & \textbf{19} & \textbf{355} & \textbf{37} & \textbf{85.4(0.71)} & \textbf{63.0(2.48)} & \textbf{0.64(0.02)}\tabularnewline
 & MV-GLMB & 556 & 524 & 132 & 57.1(3.20) & 37.9(2.88) & 0.93(0.01) & 428 & 396 & 107 & 67.0(2.77) & 44.4(2.37) & 0.91(0.01)\tabularnewline
 & MS-GLMB & 293 & 696 & 153 & 59.6(3.62) & 34.9(2.18) & 0.94(0.01) & 162 & 566 & 126 & 69.7(2.96) & 42.6(2.46) & 0.93(0.01)\tabularnewline
\hline 
\multirow{3}{*}{CMC4} & Ours & \textbf{1} & \textbf{92} & \textbf{1} & \textbf{76.7(0.51)} & \textbf{73.9(3.91)} & \textbf{0.63(0.04)} & \textbf{0} & \textbf{91} & \textbf{1} & \textbf{77.2(0.19)} & \textbf{74.2(3.82)} & \textbf{0.56(0.05)}\tabularnewline
 & MV-GLMB & 17 & 201 & 6 & 44.2(1.80) & 55.7(0.79) & 0.86(0.01) & 9 & 193 & 7 & 48.1(1.06) & 56.7(0.67) & 0.84(0.02)\tabularnewline
 & MS-GLMB & 37 & 213 & 5 & 36.5(12.69) & 50.4(7.10) & 0.87(0.02) & 29 & 205 & 6 & 40.4(11.17) & 52.3(6.52) & 0.85(0.02)\tabularnewline
\hline 
\multirow{3}{*}{CMC5} & Ours & \textbf{89} & \textbf{506} & \textbf{34} & \textbf{83.0(0.39)} & \textbf{39.1(2.41)} & \textbf{0.92(0.01)} & \textbf{29} & \textbf{446} & \textbf{33} & \textbf{86.3(0.34)} & \textbf{40.5(2.31)} & \textbf{0.91(0.01)}\tabularnewline
 & MV-GLMB & 526 & 822 & 106 & 60.8(7.72) & 21.7(3.78) & 0.96(0.01) & 331 & 627 & 99 & 71.5(7.58) & 26.7(3.26) & 0.96(0.01)\tabularnewline
 & MS-GLMB & 482 & 955 & 158 & 57.0(5.84) & 18.1(1.67) & 0.97(0.00) & 227 & 701 & 152 & 70.9(5.43) & 23.7(2.30) & 0.98(0.00)\tabularnewline
\hline 
\end{tabular}
\end{table*}

Quantitative comparison with other multi-view MOT algorithms are shown in Tab.~\ref{tbl:wildtrack} for the WT dataset, Tab. \ref{tbl:cmc} for the CMC dataset, and Tab. \ref{tbl:cmc-turnof30f} for the CMC dataset with all cameras turned off mid-scene to assess re-identification. The higher MOTA, IDF1 scores, and lower OSPA\!$^{\texttt{(2)}}$ errors, indicate superior performance of the proposed MOT filter (note that there is a very small performance difference between the two different implementations of the adaptive birth model, which will be discussed in the ablation study). For the CMC dataset, we excluded the MT, PT, and ML scores since they give no useful insight given the small number of objects in the scenes. 
\begin{figure}[h]
\centering
\includegraphics[width=0.5\textwidth]{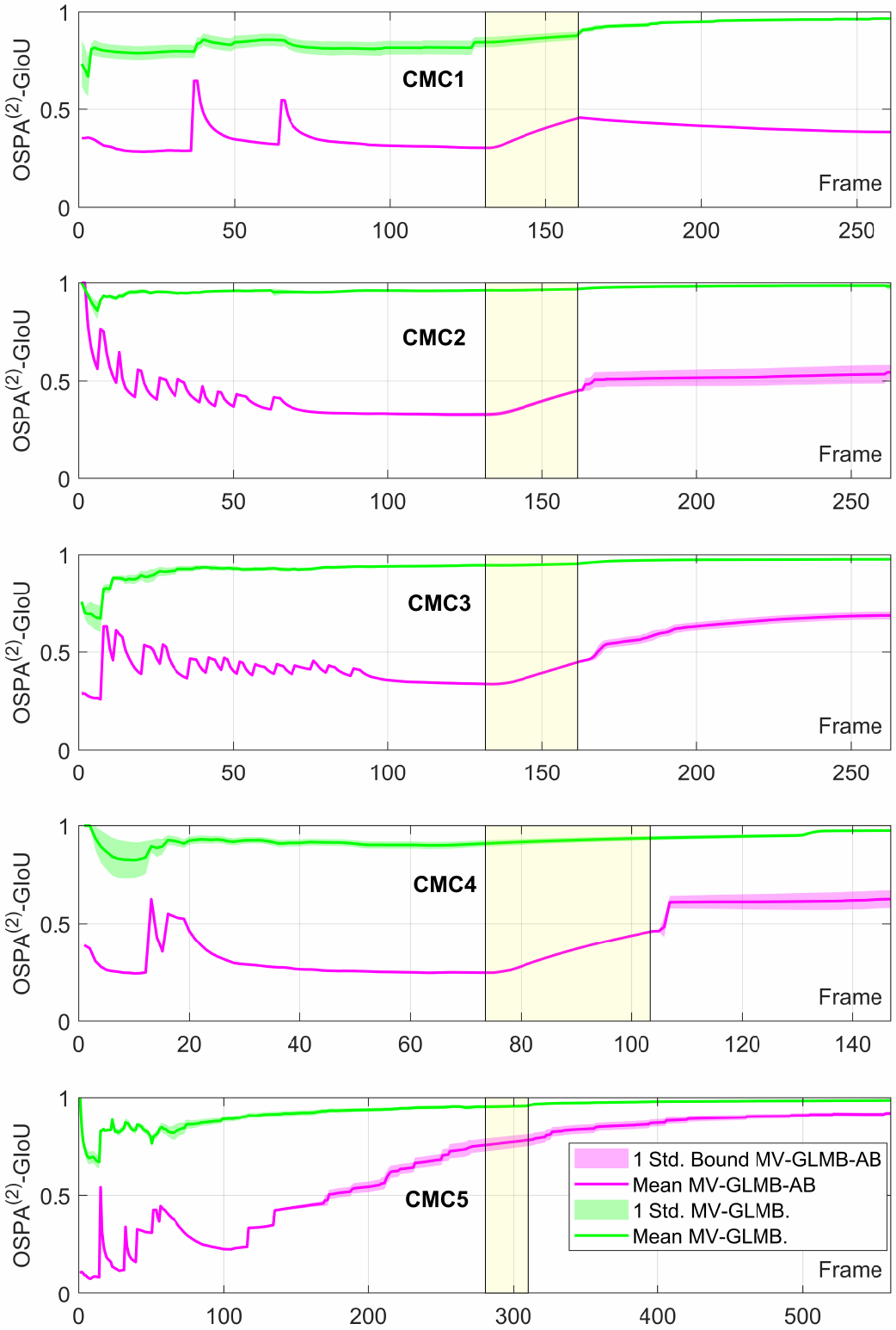}
\caption{Track re-identification: OSPA$^{(2)}$ error of the proposed filter with CSTrack detection. Object disappearance-reappearance in the CMC dataset is simulated by turning off all cameras mid-scene for 30 frames (indicated in yellow). Tracking errors for MV-GLMB almost saturate at the maximum value. Except in CMC5 where re-identification fails because the features are not stable, the proposed filter has considerably lower tracking errors at all times. \label{fig:cmcturnoff30f}}
\end{figure}
The poorer performance of the MV-GLMB and MS-GLMB filters arises from poor track initiation/re-identification and occlusion handling (both filters cannot re-identify tracks, and the MS-GLMB filter does not account for occlusions). This can be seen from the OSPA$^{(2)}$ error curves in Fig. \ref{fig:cmcturnoff30f} (the error at time \textit{k} is computed over the window from the initial time to time \textit{k}). In CMC1, the error increases when the cameras are turned off and decreases when the cameras are turned on, indicating correct re-identifion. In CMC2 and CMC4, although most tracks are correctly re-identified, the error does not decrease after the cameras are turned on because some tracks are assigned incorrect IDs. In CMC3 and CMC5, due to the high object density and severe occlusion, the features are unstable. Consequently, only a small number of tracks are recalled, and hence the OSPA$^{(2)}$ error increase.  

Tab.~\ref{tbl:runtime} shows the average run-time in FPS (frame per second, on a desktop with an Intel(R) Core(TM) i7-7700K CPU @ 4.20GHz Processor without any GPU accelerations), for each 3D MV-MOT methods in the WT and CMC datasets. The proposed filter shows improved processing speed compared to MV-GLMB \cite{ong2020bayesian}, and are able to track objects on-line. Although there are only 3 objects in CMC4, the processing time increases since we also estimate object poses. The run-time on the WT dataset is higher than that on the CMC dataset due to the higher number of objects. 

\begin{table}[!ht]
\centering
\global\long\def\arraystretch{1.3}%
 \caption{True number of objects in the sequences and MC means (1 standard deviation is shown in parenthesis) of run-time, in FPS, for different filters. The `$\ast$' indicate our filter, and the best result for each row is \textbf{Bolded}. \label{tbl:runtime}}
 \scriptsize
\begin{tabular}{|c|c|c|c|}
\hline 
Seq. & No. Obj. & MV-GLMB & MV-GLMB-AB{*}\tabularnewline
\hline 
\hline 
WT & 24 & 0.02 (0.18) & \textbf{0.06(0.01)}\tabularnewline
CMC1 & 3 & 0.62 (1.53) & \textbf{28.5(0.12)}\tabularnewline
CMC2 & 10 & 0.04 (0.41) & \textbf{7.0(0.33)}\tabularnewline
CMC3 & 15 & 0.05 (1.51) & \textbf{4.6(0.11)}\tabularnewline
CMC4 & 3 & 0.07 (0.09) & \textbf{3.6(0.42)}\tabularnewline
CMC5 & 7 & 0.02 (0.06) & \textbf{2.7(0.07)}\tabularnewline
\hline 
\end{tabular}
\end{table}

\subsubsection{Multi-Camera Reconfiguration\label{sec:multi-cam-reconf}}

\begin{figure}[ht]
\begin{centering}
\includegraphics[width=0.5\textwidth]{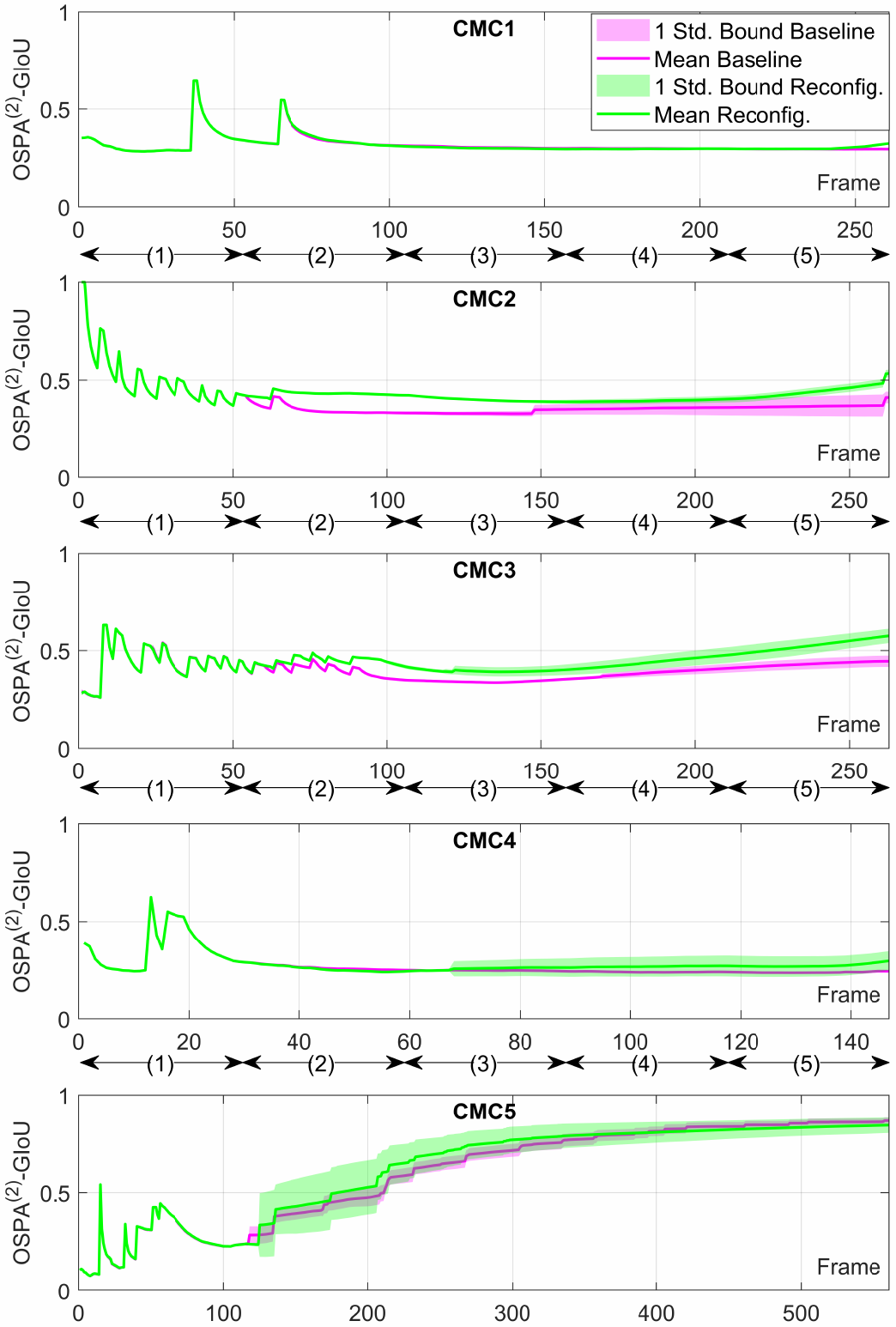}
\par\end{centering}
\caption{Multi-camera reconfiguration: OSPA$^{(2)}$ error of the proposed filter with CSTrack detection. Configuration (1): all cameras at positions 1, 2, 3, and 4 are on. Configuration (2): three cameras on at positions 2, 3, and 4. Configuration (3): three cameras on at random positions. Configuration (4): two cameras on at positions 1 and 3. Configuration (5): two cameras on at positions 2 and 4.\label{fig:configcam}}
\end{figure}

Similar to the MV-GLMB filter, our proposed filter only requires once-off training of the monocular detectors, allowing seamless operation without any interruption when cameras are added, removed, or repositioned. To demonstrate this capability, we construct, for each CMC sequence, a scenario involving five configurations over time, see Fig. \ref{fig:configcam}. The OSPA$^{(2)}$ tracking error is benchmarked against the ideal case (baseline) where all cameras are on for the entire period in Fig. \ref{fig:configcam}. 

In CMC1, the different camera configurations exhibit similar performance to the baseline due to the low object density (relative to the number of cameras). In CMC2-3, the performance degrades slightly when fewer cameras are on since it becomes harder to resolve occlusion with higher object density. Although the object density is low in CMC4, tracking upright and fallen people is more challenging due to the increased uncertainty. Hence, there is a slight performance degradation relative to the baseline. CMC5 also involves tracking upright and fallen people, but at a higher object density than CMC4. As a result the baseline and the reconfigured scenario tracking errors are similarly high due to the high object density (relative to the number of cameras). These results demonstrate that the proposed methods can adapt to camera reconfigurations on-the-fly without sacrificing the tracking performance. More details on tracking results can be found in the videos in the supplementary materials.

\subsection{Benchmarking Against Ideal Trackers\label{sec:benchmarkideal}}

In this study, we benchmark our 3D MV-MOT filter against the best possible track-by-detection performance via the combination of an ideal 3D detector with some of the best known (single-sensor) MOT algorithms. For the ideal detector we use the 3DROM detector \cite{Qiu20223DRO}, trained on 90\% of the WT dataset\footnote{Note that the WT data only provide ground truth in the ground plane, and while it is called 3DROM, this detector only provides detections in ground plane, not in full 3D.}, which is almost perfect since it is trained directly on the ground truth, and is far better than the CSTrack/FairMOT detector used for MV-MOT, see Tab. \ref{tbl:detquality-WT}. The baseline single-sensor MOT algorithms include the GLMB \cite{vo2013labeled}, MHT, JPDA, GNN filters \cite{blackman1999design}, and the KSP-ptracker \cite{chavdarova2018wildtrack} based on the DeepOcclusion detector \cite{baque2017deep}. However, since the detection is built into the tracker, we cannot evaluate the detection quality of the DeepOcclusion detector independently. The results in Tab. \ref{tbl:wildtrack_idealcomp} show that the gaps in tracking performance are not as wide as the gaps in detection performance. Keeping in mind that the 3D detection input for the single-sensor filters are effectively ground truths, it is surprising that the proposed MV-MOT filter shows comparable performance to some of the ideal filters in certain measures.%, EarlyBird \cite{teepe2024earlybird}, and MVFlow \cite{engilberge2023multi}

\begin{table}[tbh]
\centering
\global\long\def\arraystretch{1.3}%
 \caption{Detection quality for the WT dataset, `$\ast$' indicates 3D. \label{tbl:detquality-WT}}
 \scriptsize
\begin{tabular}{|l|c|c|c|c|}
\hline 
Detector & MODA$\uparrow$ & MODP$\uparrow$ & Rcll$\uparrow$ & Prcn$\uparrow$\tabularnewline
\hline 
\hline 
3DROM{*} & 93.50 & 75.90 & 96.20 & 97.20\tabularnewline
\hline 
FairMOT & 28.92 & 65.46 & 69.61 & 63.11\tabularnewline
CSTrack & 11.67 & 64.56 & 70.27 & 54.53\tabularnewline
\hline 
\end{tabular}
\end{table}

\begin{table*}[tbh]
\centering
% \global\long\def\arraystretch{1.3}%
 \caption{Tracking performance (in the \textit{ground plane}) of our filter with CSTrack detections and single-sensor filters with ideal 3D detections, on the WT dataset. The best result for each column is \textbf{Bolded}.\label{tbl:wildtrack_idealcomp}}
 \scriptsize
\begin{tabular}{|c|c|ccccccccc|}
\hline 
Detector & Tracker & MT$\uparrow$ & PT$\downarrow$ & ML$\downarrow$ & FP$\downarrow$ & FN$\downarrow$ & IDs$\downarrow$ & MOTA$\uparrow$ & IDF1$\uparrow$ & OSPA\!$^{\texttt{(2)}}$\!$\downarrow$\tabularnewline
\hline 
\hline 
CSTrack & MV-GLMB-AB{*} & 101 & 134 & 77 & 5040 & 2804 & 273 & 14.7(3.37) & 50.9(2.11) & 0.79(0.01)\tabularnewline
\hline 
FairMOT & MV-GLMB-AB{*} & 119 & 127 & 66 & 3399 & 2463 & 215 & 36.1(2.60) & 58.4(1.98) & 0.73(0.01)\tabularnewline
\hline 
\multirow{4}{*}{3DROM } & GLMB  & \textbf{167} & 107 & 39 & \textbf{136} & \textbf{1501} & 181 & \textbf{81.6} & \textbf{86.4} & \textbf{0.19}\tabularnewline
 & MHT  & 41 & 125 & 147 & 502 & 4083 & 266 & 51.0 & 50.2 & 0.31\tabularnewline
 & JPDA  & 171 & 85 & 57 & 1522 & 1770 & 368 & 63.0 & 60.1 & 0.39\tabularnewline
 & GNN  & 168 & 82 & 63 & 1911 & 1801 & 489 & 57.6 & 55.4 & 0.52\tabularnewline
\hline 
DeepOcclusion  & KSP-ptracker  & 72 & \textbf{74} & \textbf{25} & 2007 & 5830 & \textbf{103} & 72.2 & 78.4 & 0.75\tabularnewline
%\hline 
%MVFlow  & muSSP \cite{wang2019mussp}  & 38 & - & 2 & - & - & - & 91.2 & 93.4 & -\tabularnewline
%\hline 
%EarlyBird  & MOTDT  & 35 & - & 2 & - & - & - & 89.5 & 92.3 & -\tabularnewline % only track from frame 360, contain 45 unique objects
\hline 
\end{tabular}
\end{table*}

\subsection{Ablation Study\label{subsec:exp-ablation} }
\subsubsection{Sensitivity to Occlusion Model}
In this study, we assess the effect of using object features with various occlusion models, and compare the two implementations of the adaptive birth model. In particular, we compare the tracking performance of our occlusion model (IoA), the line-of-sight model (LoS) \cite{ong2020bayesian}, and the constant detection probability model, with and without object features on the CMC5 sequence. Tab.~\ref{tbl:ablation} indicates that the best performance (in IDF1 score and OSPA\!$^{\texttt{(2)}}$ error) is the combined use of object feature and the IoA occlusion model, thereby demonstrating the benefits of our proposed filter. 

\begin{table*}[tbh]
\centering
% \global\long\def\arraystretch{1.3}%
 \caption{Tracking performance for different combinations of occlusion models (Occ.) and usage/non-usage of object features (Feat.): MC means and 1 standard deviation (shown in parenthesis, only reported for the main measures). The best result for each sequence is \textbf{Bolded}.\label{tbl:ablation}}
 \scriptsize
\begin{tabular}{|P{0.6cm}|c|P{0.2cm}P{0.2cm}P{0.2cm}P{1cm}P{1.1cm}P{1.3cm}|P{0.2cm}P{0.2cm}P{0.2cm}P{1cm}P{1.1cm}P{1.3cm}|}
\hline 
\multirow{1}{*}{Occ.} & \multirow{1}{*}{Feat.} & \multicolumn{6}{c|}{Evaluation with 3D ellipsoid estimates} & \multicolumn{6}{c|}{Evaluation with ground plane estimates}\tabularnewline
\hline 
 &  & FP$\downarrow$ & FN$\downarrow$ & IDs$\downarrow$ & MOTA$\uparrow$ & IDF1$\uparrow$ & OSPA\!$^{\texttt{(2)}}$\!$\downarrow$ & FP$\downarrow$ & FN$\downarrow$ & IDs$\downarrow$ & MOTA$\uparrow$ & IDF1$\uparrow$ & OSPA\!$^{\texttt{(2)}}$\!$\downarrow$\tabularnewline
\hline 
\hline 
\multirow{2}{*}{IoA} & \multirow{1}{*}{\xmark} & \textbf{82} & \textbf{292} & 34 & \textbf{89.0(0.43)} & 45.5(3.30) & 0.92(0.01) & \textbf{31} & 240 & 31 & 91.8(0.44) & 45.9(3.30) & 0.91(0.01)\tabularnewline
 & \checkmark & 83 & 328 & \textbf{27} & 88.1(0.69) & \textbf{50.9(3.02)} & \textbf{0.87(0.02)} & 32 & 277 & \textbf{24} & 91.0(0.51) & \textbf{51.9(3.04)} & \textbf{0.86(0.02)}\tabularnewline
 & \checkmark(No Recall) & 85 & 288 & 34 & 89.0(0.52) & 45.1(2.26) & 0.92(0.01) & 33 & \textbf{236} & 30 & \textbf{91.9(0.48)} & 45.5(2.26) & 0.91(0.01)\tabularnewline
\hline 
\multirow{2}{*}{LoS} & \multirow{1}{*}{\xmark} & 88 & 301 & 32 & 88.6(0.53) & 46.6(3.13) & 0.91(0.01) & 36 & 249 & 32 & 91.5(0.45) & 47.0(3.14) & 0.91(0.01)\tabularnewline
 & \multirow{1}{*}{\checkmark} & 94 & 331 & \textbf{25} & 87.9(0.58) & 49.0(3.84) & \textbf{0.87(0.02)} & 41 & 278 & \textbf{24} & 90.8(0.44) & 50.2(3.86) & \textbf{0.86(0.02)}\tabularnewline
\hline 
\multirow{2}{*}{Const.} & \multirow{1}{*}{\xmark} & \textbf{55} & 330 & 37 & 88.6(0.43) & 41.7(3.44) & 0.93(0.01) & \textbf{23} & 298 & 34 & 90.4(0.43) & 42.3(3.34) & 0.93(0.01)\tabularnewline
 & \multirow{1}{*}{\checkmark} & 59 & 356 & 31 & 88.0(0.74) & 49.7(5.51) & 0.88(0.02) & 25 & 322 & 27 & 89.9(0.78) & 51.2(5.53) & 0.87(0.02)\tabularnewline
\hline 
\end{tabular}
\end{table*}

\begin{figure}[h]
\begin{centering}
\includegraphics[width=0.5\textwidth]{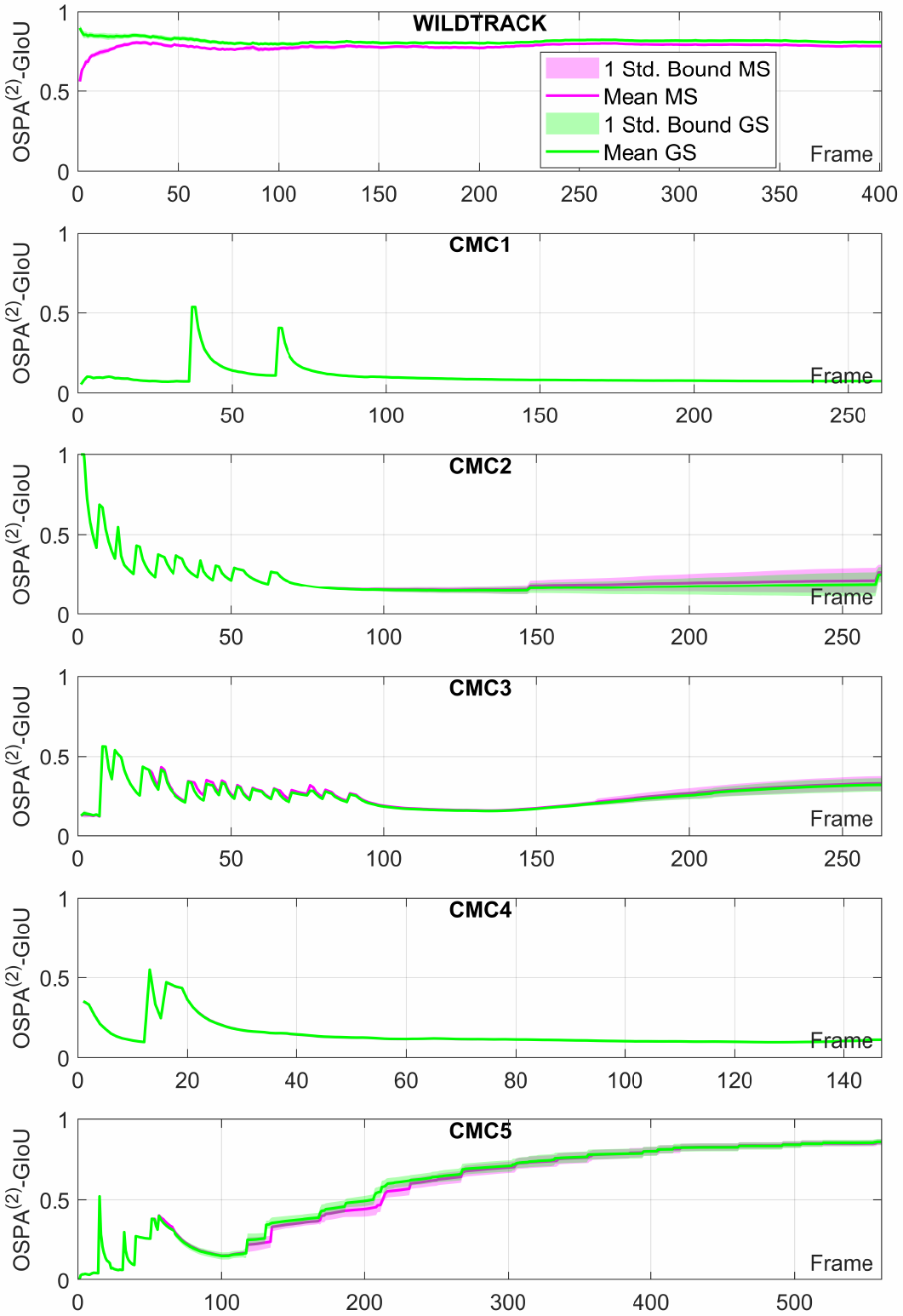}
\par\end{centering}
\caption{OSPA$^{(2)}$ tracking error (in the \textit{ground plane}) of our filter with the CSTrack detector. MS: mean-shift clustering adaptive birth implementation. Gibbs: Gibbs-sampling adaptive birth implementation. \label{fig:msga_compare}}
\end{figure}

Due to the small difference between the overall tracking performance of the mean-shift clustering (MS) and Gibbs-Sampling (GS) \cite{trezza2022multi} implementations of the adaptive birth model (of Subsection \ref{sec:adaptbirth}), to distinguish them, we need to examine their OSPA\!$^{\texttt{(2)}}$ error curves (computed as per Figs. \ref{fig:cmcturnoff30f} and \ref{fig:configcam}). Fig. \ref{fig:msga_compare}, indicates that for the WT dataset the MS implementation provides better track initialization than GS with lower OSPA\!$^{\texttt{(2)}}$ error at the beginning of the scenario. In the CMC dataset, where the area of interest is significantly smaller, the two implementations show nearly identical performance.  

\begin{table*}[tbh]
\centering{}
\global\long\def\arraystretch{1.3}%
 \caption{Main evaluation criteria for the best-hypothesis approximation and the standard trackers with CSTrack detections. Means and one standard deviations (in parenthesis) are reported for the standard tracker. Evaluation on the CMC dataset is done in 3D, and evaluation on the WT dataset is done in the ground plane. The best result for each sequence is \textbf{Bolded}.\label{tbl:ablation_hyponum}}
\scriptsize
\begin{tabular}{|c|l|ccccccc|}
\hline 
Dataset & Tracker & FP$\downarrow$ & FN$\downarrow$ & IDs$\downarrow$ & MOTA$\uparrow$ & IDF1$\uparrow$ & OSPA\!$^{\texttt{(2)}}$\!$\downarrow$ & FPS$\uparrow$\tabularnewline
\hline 
\hline 
\multirow{2}{*}{CMC1} & Single Hypothesis & \textbf{0} & \textbf{4} & 0 & \textbf{99.4} & \textbf{99.7} & \textbf{0.30} & \textbf{625.42}\tabularnewline
 & Multiple Hypotheses & \textbf{0} & \textbf{4} & 0 & \textbf{99.4(0.00)} & \textbf{99.7(0.00)} & \textbf{0.3(0.00)} & 28.5(0.12)\tabularnewline
\hline 
\multirow{2}{*}{CMC2} & Single Hypothesis & \textbf{16} & 301 & 55 & 82.1 & 38.5 & 0.91 & \textbf{112.34}\tabularnewline
 & Multiple Hypotheses & 18 & \textbf{36} & \textbf{5} & \textbf{97.1(1.43)} & \textbf{90.2(6.62)} & \textbf{0.41(0.03)} & 28.5(0.12)\tabularnewline
\hline 
\multirow{2}{*}{CMC3} & Single Hypothesis & 78 & 538 & 87 & 75.1 & 37.6 & 0.9 & \textbf{72.58}\tabularnewline
 & Multiple Hypotheses & \textbf{57} & \textbf{99} & \textbf{15} & \textbf{93.9(1.15)} & \textbf{78.6(5.03)} & \textbf{0.45(0.03)} & 7.0(0.33)\tabularnewline
\hline 
\multirow{2}{*}{CMC4} & Single Hypothesis & 3 & 10 & 1 & 96.5 & 86.6 & 0.5 & \textbf{208.63}\tabularnewline
 & Multiple Hypotheses & 0 & \textbf{9} & \textbf{0} & \textbf{97.5(0.29)} & \textbf{98.7(0.16)} & \textbf{0.24(0.00)} & 4.6(0.11)\tabularnewline
\hline 
\multirow{2}{*}{CMC5} & Single Hypothesis & 172 & 919 & 149 & 66.6 & 11.5 & 0.99 & \textbf{71.35}\tabularnewline
 & Multiple Hypotheses & \textbf{83} & \textbf{328} & \textbf{27} & \textbf{88.1(0.69)} & \textbf{50.9(3.02)} & \textbf{0.87(0.02)} & 3.6(0.42)\tabularnewline
\hline 
\multirow{2}{*}{WT} & Single Hypothesis & \textbf{1621} & 7079 & 1561 & -7.8 & 7.1 & 0.99 & \textbf{16.59}\tabularnewline
 & Multiple Hypotheses & 5040 & \textbf{2804} & \textbf{273} & \textbf{14.7(3.37)} & \textbf{50.9(2.11)} & \textbf{0.79(0.01)} & 2.7(0.07)\tabularnewline
\hline 
\end{tabular}
\end{table*}

\subsubsection{Best Hypothesis Approximation}\label{subsubsec:best-hypo}
In general, reducing the number of components (hypotheses) decreases the computation time, but at the expense of tracking performance. However, the performance degradation may not be significant in scenarios with a high signal-to-noise ratio (SNR), i.e., high detection probability and low false alarms. In this ablation study, we investigate an extreme case where we only propagate the best hypothesis in the MV-GLMB-AB filter (see Remark 2) and evaluate the performance of this approximate filter on both the CMC and WT datasets. 

Tab.~\ref{tbl:ablation_hyponum} presents tracking performance comparison for the MV-GLMB-AB filter and its single-hypothesis approximation. Observe that in the CMC1 and CMC4 sequences, the single-hypothesis MV-GLMB-AB filter is significantly faster than the MV-GLMB-AB filter without significant tracking performance degradation, due to the high SNRs. As expected, in other data sequences where the numbers of miss-detections and false alarms are high, the performance gaps are considerable. Nonetheless, the significant increase in processing speed renders the single-hypothesis MV-GLMB-AB filter suitable for real-time 3D tracking, especially with the continual improvement in detection/segmentation techniques.

Note also from Tab.~\ref{tbl:ablation_hyponum} that the single-hypothesis MV-GLMB-AB filter yields significant increases in ID switches. This is because tracks that are discarded along with the non-optimal hypotheses cannot be recalled later when evidence supporting their existence accumulates. As a result, the filter incorrectly initiates new tracks, leading to a high number of ID switches. This can be improved via an ad-hoc scheme that retains significant tracks from discarded hypotheses and recalls them later when there is sufficient evidence supporting their existence. Reducing ID switches in a principled manner requires further investigation.

\section{Conclusion\label{sec:conclusion} }

We have exploited recent advancements in 2D detection and multi-view fusion to develop a 3D MV-MOT filter that processes 2D detections from monocular cameras, which avoids expensive 3D object detector training. The proposed MV-MOT filter integrates automatic track initialization, re-identification, occlusion handling, and data association into a single Bayesian filtering framework while at the same time taking advantage of object features to improve efficiency. Performance evaluation on challenging scenarios demonstrated significant improvements of the proposed filter over existing MV-MOT solutions. Ablation studies also show its robustness when camera configurations are changed on-the-fly, and the advantages of the proposed occlusion and adaptive birth models to resolve occlusions and automatically initiates/re-identifies tracks. To the best of our knowledge, the proposed filter is the first to perform track re-identification in 3D from 2D detections.  Re-identification could be improved using features that are unique to the objects and time-invariant (or vary slowly with time), which is still an open topic in computer vision.

\bibliographystyle{IEEEtran}
\bibliography{reflib_abbr}

\end{document}